\documentclass[times, preprint, 10pt]{elsarticle}

\usepackage{xcolor}
\usepackage{color}
\usepackage{amssymb}
\usepackage{amsmath}
\journal{Pattern Recognition}

\begin{document}

\begin{frontmatter}



\title{Kyrtos: A Methodology for Automatic Deep Analysis of Graphic Charts with Curves in Technical Documents}

\author[label1]{Michail S. Alexiou\footnote[1]{Corresponding author. E-mail address: malexiou@kennesaw.edu}}
\author[label2]{Nikolaos G. Bourbakis}
\affiliation[label1]{organization={Kennesaw State University},
            addressline={Technology Pkwy SE},
            city={Marietta},
            postcode={30060},
            state={Georgia},
            country={USA}}

\affiliation[label2]{organization={Wright State University},
            addressline={3640 Colonel Glenn Hwy},
            city={Dayton},
            postcode={45435},
            state={Ohio},
            country={USA}}

%% Abstract
\begin{abstract}
Deep Understanding of Technical Documents (DUTD) has become a very attractive field with great potential due to large amounts of accumulated documents and the valuable knowledge contained in them. In addition, the holistic understanding of technical documents depends on the accurate analysis of its particular modalities, such as graphics, tables, diagrams, text, etc. and their associations. In this paper, we introduce the Kyrtos methodology for the automatic recognition and analysis of charts with curves in graphics images of technical documents. The recognition processing part adopts a clustering based approach to recognize middle-points that delimit the line-segments that construct the illustrated curves. The analysis processing part parses the extracted line-segments of curves to capture behavioral features such as direction, trend and etc. These associations assist the conversion of recognized segments’ relations into attributed graphs, for the preservation of the curves’ structural characteristics. The graph relations are also are expressed into natural language (NL) text sentences, enriching the document's text and facilitating their conversion into Stochastic Petri-net (SPN) graphs, which depict the internal functionality represented in the chart image. Extensive evaluation results demonstrate the accuracy of Kyrtos' recognition and analysis methods by measuring the structural similarity between input chart curves and the approximations generated by Kyrtos for charts with multiple functions.
\end{abstract}

% \begin{highlights}
% \item A hybrid methodology is proposed for the recognition and recoloring of monochromatic chart images containing solid or dashed curves.
% \item The methodology consists of an unsupervised learning keypoint-extraction module in conjunction with probabilistic modeling for uncertainty.
% \item Extensive experiments using a synthetic dataset show the effectiveness of the proposed method despite the increasing number of identically colored curves.
% \item The synthetic dataset is made available to the public for evaluation and comparison with future techniques.
% \end{highlights}

%% Keywords

\end{frontmatter}



\section{Introduction}
\label{intro}
%% Labels are used to cross-reference an item using \ref command.

The recent rise of published research papers has escalated the need for automated ways to extract and analyze their illustrated knowledge, thus, raising the issue of automatic document reverse engineering into a major research area \cite{doc_assist}. Document analysis is a field of pattern recognition dedicated to the automated extraction of information from either digital or handwritten documents \cite{biblio_survey}. It has been actively studied over the years and has significant applications across multiple research areas, including content-based document retrieval \cite{bunke11}, segmentation \cite{gordo13}, summarization and understanding \cite{layoutllm}. Document summarization methodologies focus on selecting representative words and sentences from paragraphs of text, using clustering-based methods \cite{selosse} to guide their decision-making process. Document segmentation techniques recognize text and non-text regions \cite{tanlu21, riba22}, using OCR methods and transforming the problem space into that of structural graph recognition.

Similarly, text-based document retrieval deals with extracting embeddings from the available document text \cite{passalis18}, which are used to evaluate their similarity score against other documents. Text-based document retrieval evolved into content-based, to overcome the former’s lack of holistic knowledge representation. Content-based document retrieval methods focus on the recognition of the document’s layout (i.e., location of text in correspondence to images) and structural characteristics of its images (i.e., shape, color, etc.), which are converted into fuzzy attributed graphs \cite{chaieb17} for graph matching-based retrieval. While the aforementioned methods constitute an integral part of automated document analysis, they focus only on a portion of the document’s deeper semantic meaning and internal knowledge, which creates the need for more granular approaches.

In order to accurately understand the information contained in a technical document \cite{cao23}, we must first dissect it into its individual modalities (i.e., tables, diagrams, charts, NL text, etc.) and associate them to capture the deep semantic meaning contained in it \cite{unifyingmodal, cvprli24}. Each modality offers a different view from the overall knowledge, about the functionality and the structure of the described methodologies. In the current research, we are focusing on the automatic analysis of chart image in technical documents. Existing techniques recognize information of charts, restricted to specific heuristic rules such as a width of one pixel \cite{nair}, or simplistic bar charts \cite{chester}. Additionally, there have been methods developed for curve analysis based on trend segmentation \cite{wupeng1}, however, they assume that the chart images have already been processed and converted into data series by external tools. A detailed analysis of the aforementioned methodologies is conducted in \cite{graph_surv}, which proves the need for an accurate and generalized chart recognition and analysis methodology with deep understanding capabilities. The goal of this research is to capture the visual knowledge illustrated in the chart images of technical documents, deduce their internal structural and behavioral associations, and represent them into a common form, like the SPNs, to improve document analysis methodologies that require deep document understanding.

Thus, in this paper we describe the novel methodology Kyrtos, for automatic recognition and understanding of the information that is illustrated in chart images containing solid or partitioned curves of different colors. We define a chart figure as the visual representation of the variations in the values of one or multiple variables with respect to the progression of time \cite{graph_surv}. These variables are represented as either curves or bars within the chart. Kyrtos is invariant to heuristic rules about width, coloring, direction and expansion of chart curves, since it is based on a selection of representative pixel-level features to approximate their structural and behavioral information. While the overall methodology was initially introduced in \cite{glossa}, focusing on the conversion of text in technical documents into SPN graphs, there remains a need for domain-specific techniques to accurately retrieve and integrate both the visible and semantic associations depicted in each modality, mimicking human perceptions. Additionally, the benefits of using attributed graphs for the representation of visual information extracted from engineering diagrams \cite{bunke11} or \cite{pinakas} tables have been studied before, but they haven’t been adapted for the representation of curves yet, to the best of our knowledge.

In summary, the main contributions of this paper are: (1) An automated method for recognizing and analyzing graphics in technical documents, overcoming structural restrictions and analyzing curves regardless of their characteristics; (2) Kyrtos integrates unsupervised learning with formal and statistical logic to improve analysis accuracy, providing a deeper understanding of chart information; (3) The introduction of the Kyrtos formal language, bridging retrieved relations and their mapping into SPNs; and (4) The conversion of chart associations into SPN format for enhanced understanding and recreation of chart functionality. The paper is organized as follows: Section \ref{sec_relatedwork} reviews relevant literature, Section \ref{sec_method} outlines the proposed methodology, Section \ref{sec_chartreco} details the chart recognition method, Section \ref{sec_formallang} discusses the Kyrtos formal language, Section \ref{sec_chartanalysis} covers the chart analysis method and transformation into SPNs, Section \ref{sec_eval} presents evaluation results, and Section \ref{sec_conclude} offers conclusions and future research directions.

\section{Related Works}\label{sec_relatedwork}

Literature reviews studying the topics of graphics detection, recognition and understanding methodologies have already been conducted extensively by Davila et. al. \cite{chart_surv} and \cite{graph_surv}. However, this section highlights notables works that have influenced the Kyrtos methodology. Lu et al. \cite{kataria1} analyze curves with highlighted data points by using a K-Median filter to discard connecting curves and reconstruct them based on data point locations, also adopted by \cite{kataria2}. Nair et al. \cite{nair} recognize the overall direction of curves, including solid and partitioned lines, using slope gradients, but limit their analysis to one-pixel-wide lines. Cliche et al. \cite{cliche} propose reverse engineering scatter plots to extract original table data by clustering scatter points with DBSCAN to determine axis ticks, associating data points with axis values based on tick locations.

Chester et al. \cite{chester} focus on recognizing bar and pie charts by grayscaling images and identifying connected components as straight line-segments for bars or circular regions for pie charts. Al-Zaidy et al. \cite{alzaidy} introduce a machine learning algorithm to detect structural chart components like axis labels, values, and legends. Al-Kady et al. \cite{alkady} propose "Chartlytics", a framework enhancing chart accessibility for visually impaired users, involving chart type classification with MobileNet, component detection via YOLO, and data extraction for generating textual descriptions. Dou et al. \cite{dou24} utilize textual documentation of vector graphics to build multi-graphs for modeling structural and spatial information, using a dual-stream graph neural network to detect and classify objects directly from vector graphics. The authors of \cite{carberry, burns} use a Bayesian inference system to recognize trends and semantics in bar graphics, but this requires prior conversion of data into XML format. Greenbacker et al. \cite{greenbacker} apply a similar inference system to identify trends in single-curve graphics, using a Support Vector Machine (SVM) to segment the curve into line segments and determine the overall trend.

Alexiou et al. \cite{alexiou_chart} concentrate on the automated analysis of bar chart images. Their approach uses deep learning to identify extract representative keypoints from the bars, which are then converted into curves and analyzed using Stochastic Petri-net (SPN) graphs to determine their behavior. Cheng et al. \cite{chart_reader} introduced ChartReader, a framework that combines chart derendering and comprehension by employing transformer architectures for component detection and extending vision-language models for question-answering (QA). To overcome the limitations of previous work and enhance visual QA for charts, Li et al. \cite{chart_vqa} suggest leveraging Large Language Models (LLMs) to decompose complex question-answer pairs, resulting in better performance.

Wu et al. \cite{wupeng1} propose a methodology for automatically recognizing the message of single-line information graphics using an SVM model to segment lines based on trends. A Suggestion Generation Module processes these segments with the chart’s XML data to generate potential high-level messages, which are then evaluated by a Bayesian Network \cite{carberry} to determine the most likely overall message. In \cite{wupeng2}, the authors detail the SVM model’s training parameters and selected features, demonstrating its effectiveness on single-line information graphics with arbitrary directions. Kim et al. \cite{kim4} use a multimodal neural net for classifying graphics images, combining a Convolutional Neural Network (CNN) for pixel data with a bag-of-words model for text information. Li et al. \cite{li23} focus on control chart pattern recognition (CCPR) in manufacturing, using Multi-Delay Weighted Ordinal Pattern (MDWOP) for feature extraction and an ensemble classifier to improve anomaly detection accuracy.

\section{The Overall Methodology} \label{sec_method}

\begin{figure}[t]%% placement specifier
\centering%% For centre alignment of image.
\includegraphics[width=\textwidth,height=5cm]{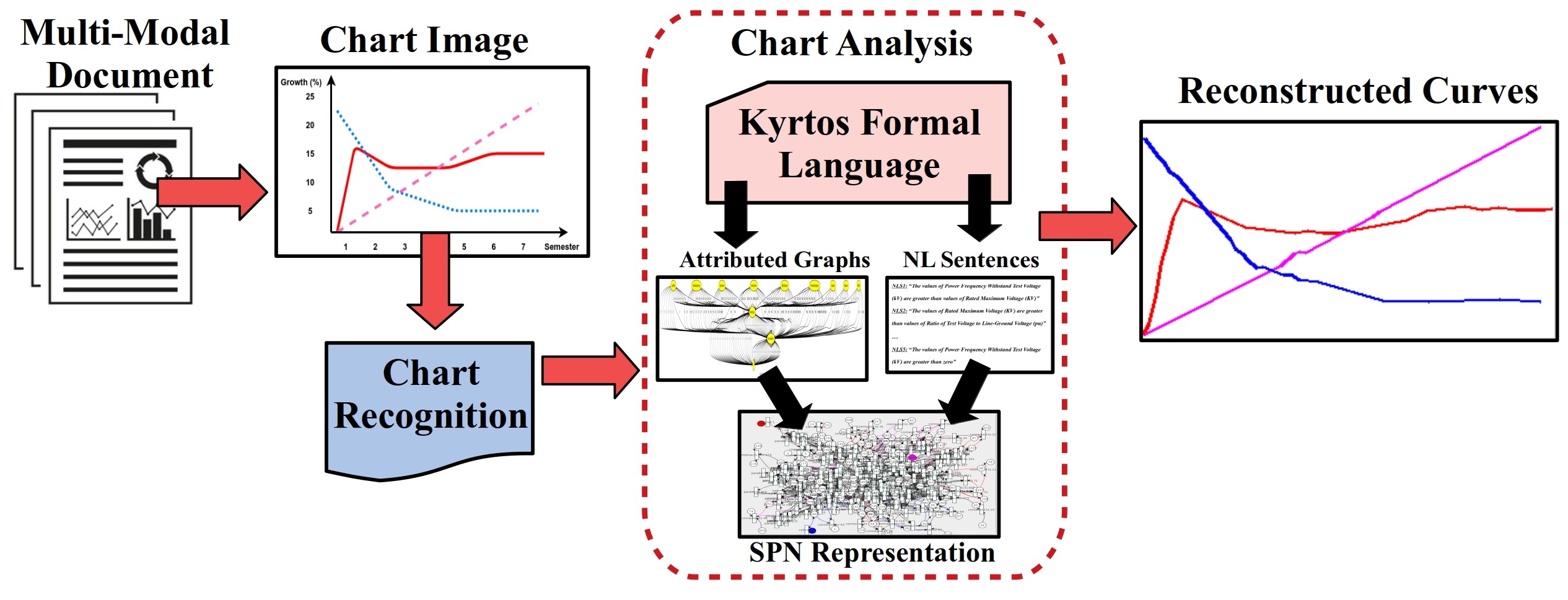}
%% Use \caption command for figure caption and label.
\caption{Pipeline of the Kyrtos chart analysis methodology.}\label{fig1}
\end{figure}

As depicted in Fig. \ref{fig1}, the proposed methodology comprises two main processing modules: (i) a chart recognition module that locates representative line-segments for each curve in a given chart image, and (ii) a chart analysis module that utilizes the Kyrtos formal language to identify behavioral patterns in the curves and convert these into SPN graphs. The chart recognition module focuses on identifying straight line-segments that depict the structure and dynamics of the curves within chart images. After extracting chart images from document pages, the recognition module processes these images to locate representative keypoints for each curve using a combination of unsupervised clustering techniques. These keypoints are the two-dimensional (2D) middle-points where each curve's direction changes. These 2D keypoints facilitate the generation of straight line-segments that form the basis for analyzing curve behavior. The generated line-segments then serve as inputs to the chart analysis module.

In the chart analysis phase, the module retrieves deeper associations among the straight line-segments both within each curve and across curves using the Kyrtos formal language. This analysis identifies potential behavioral patterns, such as parallelism and intersection among the segments, which are leveraged for the detection of high-level behaviors like growth over time. The information deduced from this analysis are then transformed into two formats: attributed graphs and natural language sentences, both illustrating the structural characteristics of the curves and their interrelations. Specifically, the natural language sentences are utilized to create SPN graphs, which delineate functional characteristics of the curves and offer a deeper comprehension of the retrieved knowledge. This is achieved once again through the Kyrtos formal language, which assists in mapping these identified patterns and segment associations into SPN kernels.  The generated natural language sentences, attributed graphs, and SPN graphs constitute the initial output of our methodology. We then use the SPN graphs to regenerate the recognized curves, which are subsequently evaluated against the original chart image using the structural similarity index to verify the accuracy of our approach.

\begin{figure}[t]%% placement specifier
\centering%% For centre alignment of image.
\includegraphics[width=0.8\textwidth,height=5.3cm]{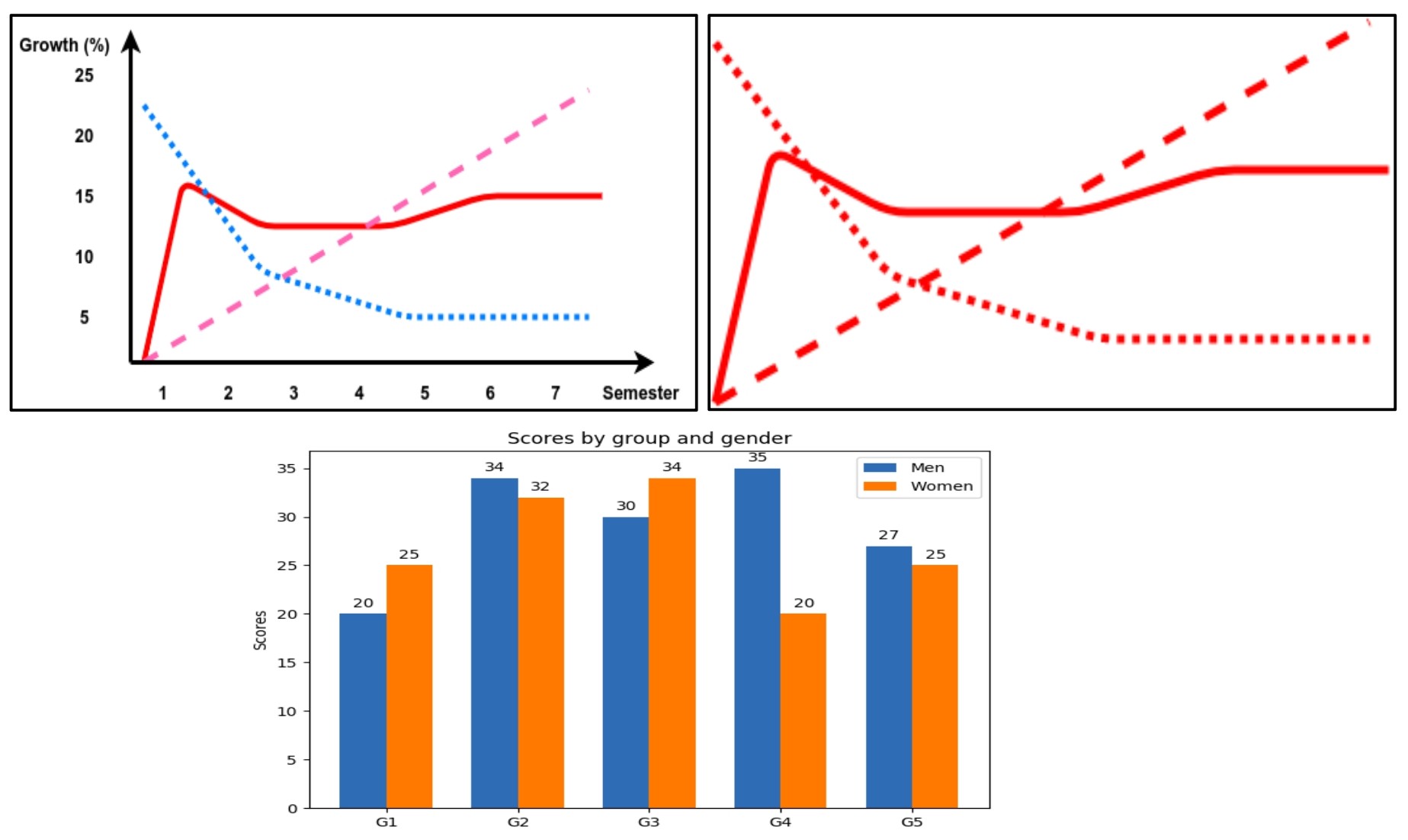}
%% Use \caption command for figure caption and label.
\caption{Mixed curves of different colors (left), partitioned curves of the same color (right) and bars chart (bottom).}\label{fig2}
\end{figure}

Chart analysis encompasses a diverse range of chart types, each requiring tailored methods to accurately interpret and approximate their behavior. This study identifies four main types of graphical figures: 1) solid or partitioned curves in different colors, 2) partitioned curves of the same color with varying shapes, 3) bar graphics, and 4) pie charts, as illustrated in Fig. \ref{fig2}. Our focus in this paper is on the first type, serving as a proof of concept for the proposed graphics understanding methodology. While global methods may be applicable for other chart processing tasks that rely on visual cues and textual data (e.g., chart summarization), chart analysis, particularly for pie charts \cite{chart_surv}, bar charts \cite{alexiou_chart}, and line charts \cite{nair}, requires specific processing methods to accurately interpret the illustrated information. In particular, the analysis of line charts with curves of varying colors presents several challenges, including accurately approximating the behavior of multiple curves regardless of their width, direction, or color. Our proposed methodology focuses on addressing these challenges by developing a robust approach for the automated analysis of charts featuring solid and partitioned curves in different colors. Throughout the paper, we use chart type 1 from Fig. \ref{fig2}, which contains a mix of continuous and partitioned lines, as a running example to illustrate the chart recognition and analysis methodologies.

\section{Recognition of Charts} \label{sec_chartreco}

\begin{figure}[t]%% placement specifier
\centering%% For centre alignment of image.
\includegraphics[width=\textwidth,height=3.2cm]{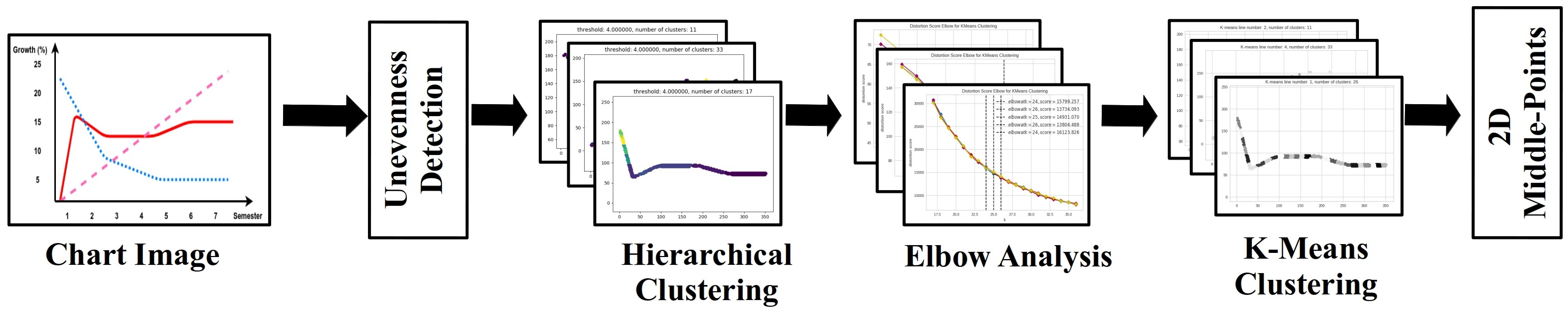}
%% Use \caption command for figure caption and label.
\caption{The Chart Recognition Methodology of Kyrtos.}\label{fig3}
\end{figure}

The methodologies presented in this paper focus on recognizing and understanding chart images by emulating human perception of graphical information. The goal is to convert different document modalities (e.g., charts, tables, text) into a unified representation for deeper knowledge extraction. Fig. \ref{fig3} illustrates the steps in our graphics recognition process. Building on prior work \cite{nair, kataria1} that used straight line-segments for one-pixel-wide curves, our approach introduces the concept of unevenness points and clustering to handle curves of varying widths. After detecting these points, iterative clustering filters them, and the midpoint of each cluster is used to construct line segments. These segments facilitate automatic recognition of parallelisms and intersections, enhancing overall understanding accuracy by merging unnecessary segments.

\subsection{Graphics Preprocessing}

In order to recognize the curves with high accuracy, we separate the part of the image containing the actual curves from the axis information. Standard image processing techniques such as segmentation and binarization, when applied in chart images cause miscoloring, missing edges or loss of entire curves. Therefore, we employ a combination of contrast and sharpen filters with the Hue-Saturation-Value (HSV) color space, which we refer to as the HSV homogeneity filter. This filter enables the recognition of any curve regardless of its color, by replacing different shades of the same color with that colors’ dominant value from the chart. We successfully distinguish the chart’s axis, grid and curves from each other, by grouping all pixels based on their respective dominant values resulting from the HSV homogeneity filter. The image is converted back to RGB colorspace and the curve recognition commences. PyTesseract OCR is used to detect the axis values and labels in the original image. The centers of axis elements’ bounding boxes are also used during the association of the information extracted from the subimage with the axis elements in the original image.

% \begin{figure}[t]%% placement specifier
% \centering%% For centre alignment of image.
% \includegraphics[width=0.75\textwidth,height=3.4cm]{figs/fig5.png}
% %% Use \caption command for figure caption and label.
% \caption{The results of transforming the colorspace of a chart to HSV, applying the HSV homogeneity filter and then transforming back to RGB.}\label{fig5}
% \end{figure}

\subsection{Detection of unevenness Points}
In order to recognize a curve’s structural and behavior characteristics, we first dissect it into the straight-lines segments that construct it. We define these straight line-segments as the segments between points in the curves where their direction changes (an unevenness occurs). After the HSV homogeneity filter has been applied to the chart image, the pixel trail of each curve is both traceable and distinguishable by a unique color. We adopt the unevenness detection methodology presented in \cite{unevennes}, where a formal solution is discussed for the issue of line recognition with randomly occurring anomalies. These anomalies are referred to as unevenness. More specifically, they define unevenness $U$ for line-segment $SL$ as the number $i$ of consecutive pixels with $i \in N$, that follow a different direction $d_i$, from the main direction $d_{main}$ of the line-segment $SL$. They develop and describe two interconnected equations in order to tackle this issue. Eq. \ref{eqn:eq1} describes the window that covers the line-segment, which contains the unevenness. $W_U$ represents the area of the window, $H_U$ represents the maximum accepted unevenness and $L_U$ represents the length of the line-segment containing the unevenness. Eq. \ref{eqn:eq2} defines the unevenness criteria as the division between the height of the window covering the line-segment, which contains the unevenness ($W_U$), and that segment’s length. The smaller the value of the unevenness criteria, the smaller the acceptance limit for the unevenness. Therefore we adjust its value automatically by assign to the variables $H_U$ and $L_U$ the values of the extracted subimage’s height and width respectively.

\begin{equation}\label{eqn:eq1}
W_U = H_U \times L_U
\end{equation}
\begin{eqnarray}\label{eqn:eq2}
e = H_U/L_U
\end{eqnarray}
\begin{equation}\label{eqn:eq3}
\lambda = (y_1 - y_0) / (x_1 - x_0)
\end{equation}

\begin{figure}[t]%% placement specifier
\centering%% For centre alignment of image.
\includegraphics[width=\textwidth,height=4.2cm]{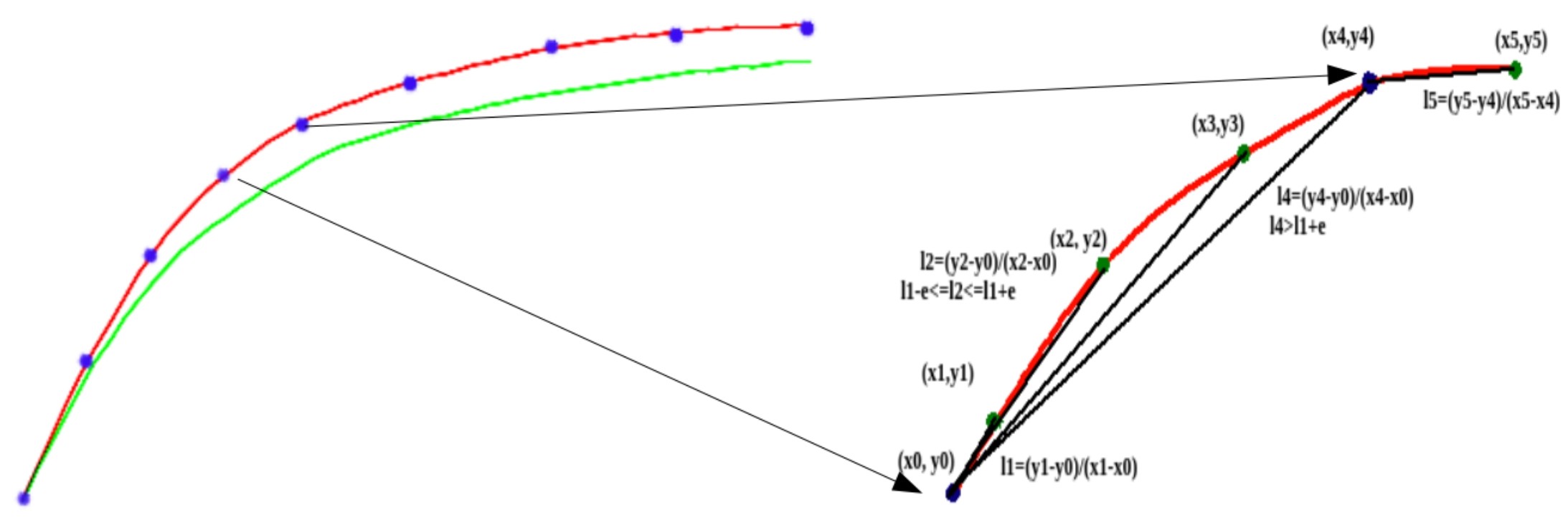}
%% Use \caption command for figure caption and label.
\caption{Example of recognizing changes in direction (unevennesses) using the slope equation and the unevenness criteria component.}\label{fig6}
\end{figure}

The equation for slope $\lambda$ between 2 points $(x_0, y_0)$ and $(x_1, y_1)$ is illustrated in \ref{eqn:eq3}. A slope limit threshold plays a crucial role in detecting changes in the direction of the curve and, thus, identifying the straight line-segment. Its value is determined by calculating the difference between the current slope value and the previous slope value. If the computed slope value for the current pixel is smaller than the previous slope value (which was obtained when the last direction change was detected) minus the threshold value for unevenness criteria threshold, it indicates a new change in direction. Similarly, if the computed slope is larger than the previous slope value plus $e$, it also signifies a change in direction. An example of the aforementioned procedure is illustrated in Fig. \ref{fig6}. The position of the pixel where the direction change was detected is selected as the unevenness point. The new slope limit is updated with the value that was calculated for the new unevenness point. This procedure continues until the entire curve trail has been parsed and it is repeated for the remainder of the curves.

\subsection{Clustering of Unevenness Points}
The detected unevenness points are sufficient for the generation of straight line-segments from curves with width of 1 or 2 pixels, as in \cite{nair}. However in case of thicker lines, the unevenness detection method proposes an arbitrary number of candidate unevenness points for the same regions in the curve. We address this issue by applying a clustering technique following the unevenness analysis, effectively filtering out any redundant 2D points identified during curve parsing. More specifically, the 2D middle-point from each cluster of unevenness points is selected to generate the straight line-segments instead. A cluster’s middle-point is different from its centroid, since the middle-point is located in the middle of the cluster, with respect to the x-axis locations of the rest its elements. The application of the unevenness criteria analysis followed by clustering enables the reservation of the curve’s direction and disregards any minor anomalies in the curve.

\begin{figure}[t]%% placement specifier
\centering%% For centre alignment of image.
\includegraphics[width=0.75\textwidth,height=5cm]{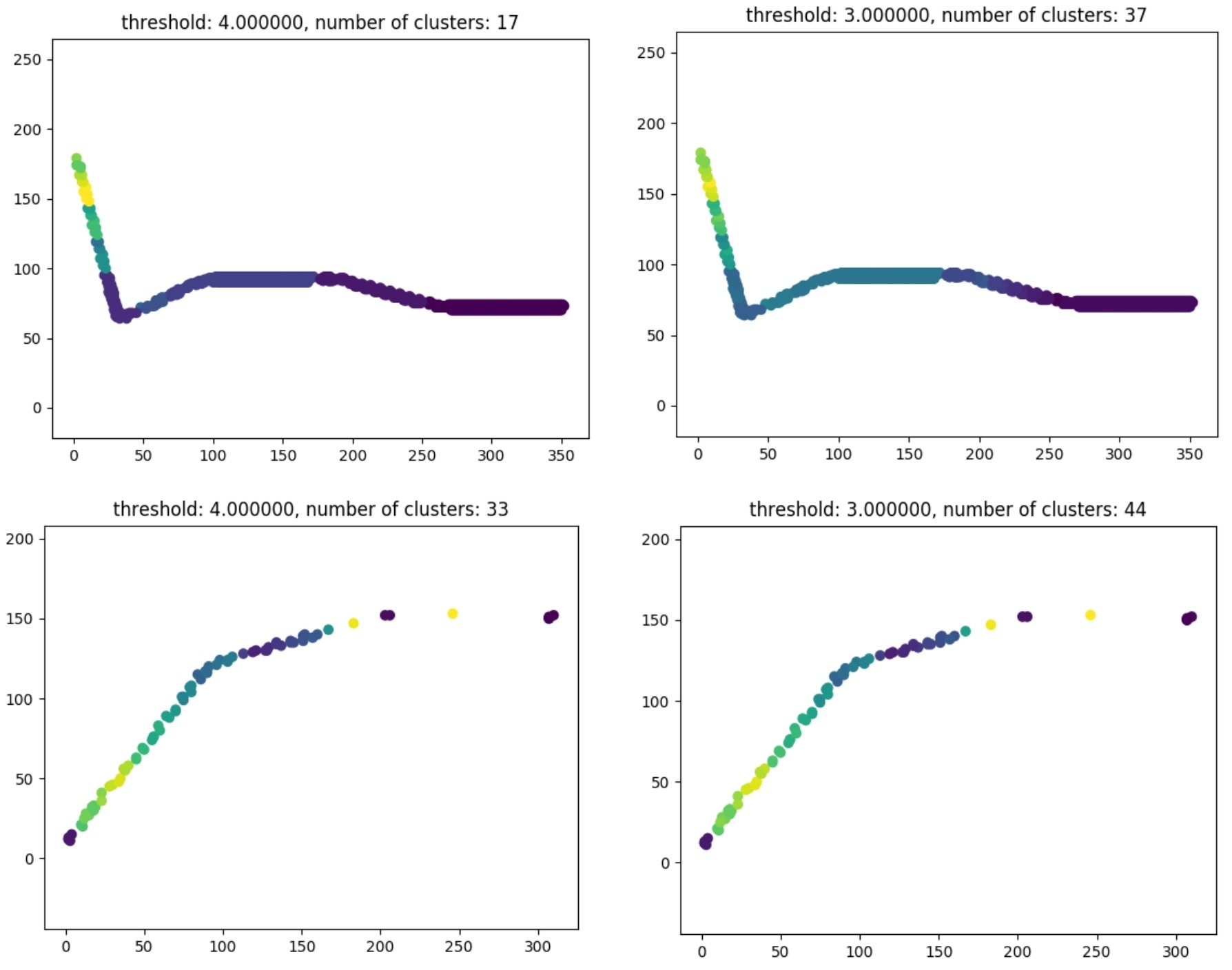}
%% Use \caption command for figure caption and label.
\caption{Example of clustering detected unevenness points for threshold of 4 pixel distance (left column) and of 3 pixel distance (right column).}\label{fig7}
\end{figure}

Predefining the number of clusters heuristically is unscalable and inaccurate due to unpredictable and varying curve thickness. Instead, we use hierarchical clustering to set boundaries for a K-Means algorithm. Hierarchical clustering groups data points based on distances, using upper and lower pixel distance thresholds to limit the search space and filter out excess unevenness points. The thresholds, determined through trial and error, yield detailed clusters at 3 pixels (upper limit) and good clusters at 4 (lower limit) pixels. Two instances of hierarchical clustering provide upper and lower cluster limits, which guide iterative K-Means clustering for each curve. The optimal cluster count is selected using the elbow technique, with K-Means run five times per iteration and the result with the lowest distortion score is maintained for the final evaluation, to overcome errors produced by different centroid starting positions. This empirical solution is based on the notion that a lower distortion score corresponds to higher number of clusters and, thus, higher detail of clustering.

\begin{figure}[t]%% placement specifier
\centering%% For centre alignment of image.
\includegraphics[width=0.9\textwidth,height=6cm]{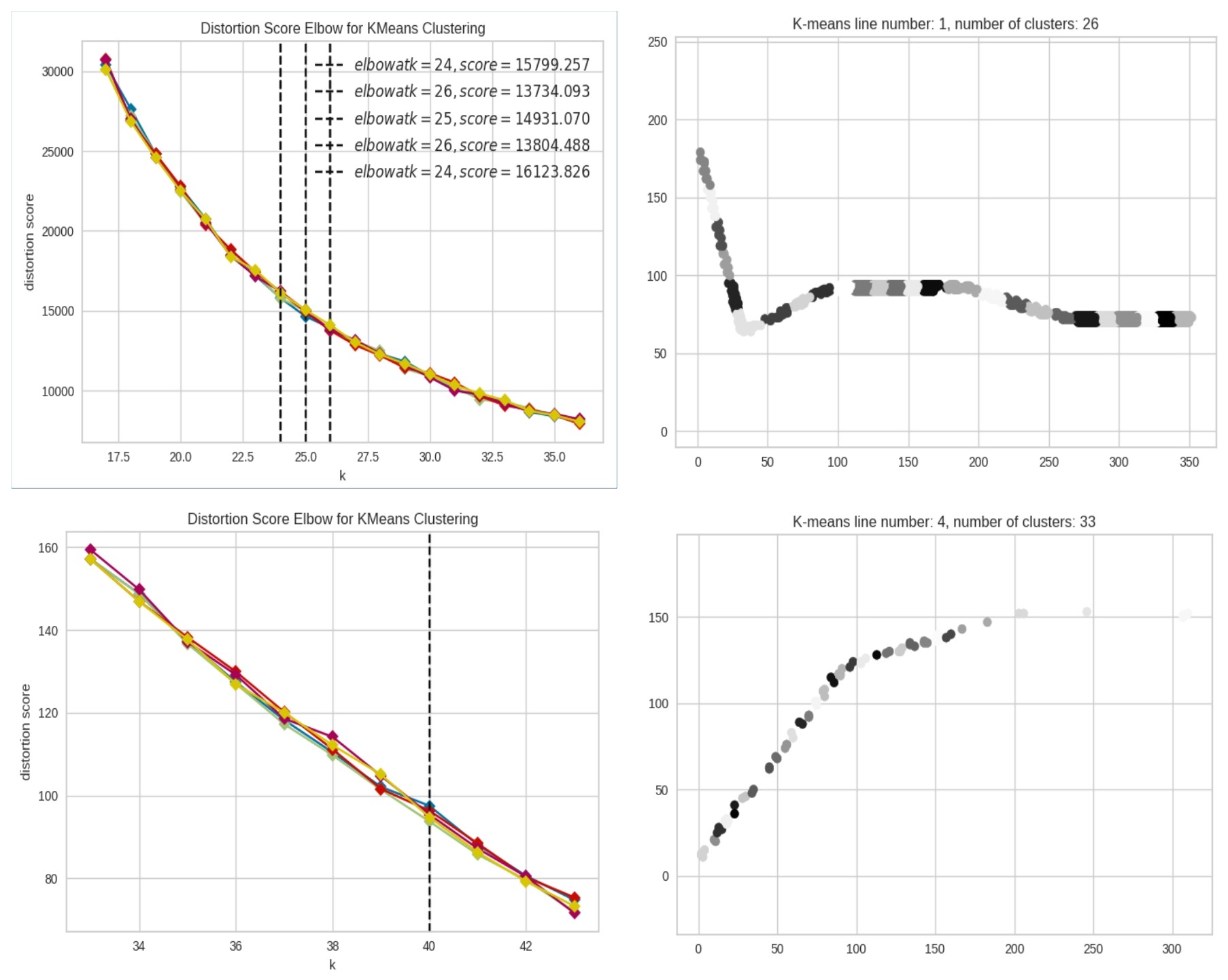}
%% Use \caption command for figure caption and label.
\caption{Example of the elbow methodology (left column) and K-means clustering 2D middle-points using the elbow results (right column).}\label{fig8}
\end{figure}

Finally, we perform K-Means clustering on the unevenness data points using the optimal cluster number that are deduced for each curve. The position of the middle-point of each cluster is calculated to produce the line-segments. The hierarchical clustering results for the running example about the red and blue curves from the type 1 chart image of Fig. 2, is presented in Fig. 7. Similarly, the results for the K-Means clustering are illustrated in Fig. 10. In particular, the right column of Fig. 8 illustrates the corresponding elbow detection results and the left column illustrates the corresponding K-Means clustering results for the red and the blue curves (curves 1 and 3) respectively. In case no optimal cluster number can be deduced during the iterative K-Means procedure, the lower boundary from the hierarchical clustering phase, is used as the cluster number for the final K-Means algorithm.

\subsection{Noise Pixels Detection}
Pixels sharing the same colors as the curves but with unrelated positions with respect to them within the image plane are considered noise. Such pixels form clusters of their own during the curve segmentation phase and their noise is propagated to the step of straight line-segment generation. Potential noise pixel or a group of pixels are distinguished from pixels of curves using a using a distance threshold. The purpose of this threshold is to eliminate pixels that may have been mistakenly identified as part of a specific curve. Specifically, its value is the same as the identified the width of the curve itself. If the distance between a middle-point pixel and its closest pixel on the curve exceeds this threshold, it is considered as noise and gets disregarded.

\section{The Kyrtos Formal Language} \label{sec_formallang}
The Kyrtos formal language is engineered to capture and represent the semantic and syntactic attributes of line segments in chart images, facilitating their mapping into SPN kernels. This language plays a critical role in two main steps of the methodology: (i) identifying behavioral patterns from visual data and (ii) translating these patterns into structured SPN graphs for further analysis. Formal languages enable precise definition and systematic recognition of complex visual elements, providing a robust framework for interpreting diverse visual information. In this paper, Kyrtos extends the principles of the Glossa formal language \cite{glossa}, which is used to formalize kernels from natural language sentences extracted from documents.

\subsection{Alphabet}
We begin with the definition of the alphabet for the Kyrtos language. In order to achieve this we must consider each curve of the graphics image as its own entity. Each curve consists of straight line-segments that connect with each other for an angle of specific degrees. So $C_i = \{x \mid x \text{ is the } i\text{th curve of the input graphics image } GR\}$. Additionally, $SL_{ij}$ = \{$x \mid x$ is a straight line-segment that constitutes part of the curve $C_i$ from the input graphics image $GR$\}, since $SL_{ij}$ now represents the $j$th straight line-segment for the $i$th curve of the graphics image. Finally, $Q_{ijkn}$ = \{$x \mid x$ is the associations of among $j$th and $n$th  straight line-segments of the $i$th curve and the $k$th curves respectively from the input graphics image $GR$\}. These associations include connections, intersections and parallelism. The $Q$ symbol is used in an effort to formulate the relationships both among different line-segments of the same curve and among straight line-segments of different curves.

\begin{figure}[t]%% placement specifier
\centering%% For centre alignment of image.
\includegraphics[width=0.8\textwidth,height=3.4cm]{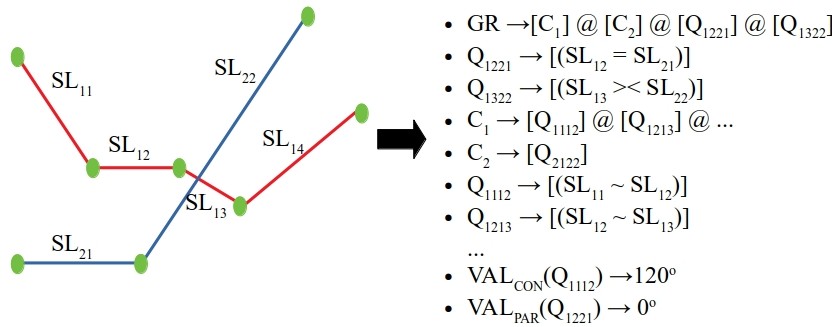}
%% Use \caption command for figure caption and label.
\caption{Formulation Effort Example for Simple Curves.}\label{fig9}
\end{figure}

We illustrate an example of formulation in Fig. \ref{fig9}. This simple graphics image $GR$ is described by the structural information of its curves $C_1$ and $C_2$, as well as their associations with each other. These associations are represented in the form of intersection and parallelism relations among their corresponding straight line-segments. For example,  $"SL_{13} >< SL_{22}"$ corresponds to the intersection among straight line-segments $SL_{13}$ and $SL_{22}$ from curves $C_1$ and $C_2$ respectively. Furthermore, $"SL_{11} \sim SL_{12}"$ illustrates the connection formed among segments $SL_{11}$ and $SL_{12}$ from $C_1$. Finally, $VAL{CON}()$ represents the value of the particular connection which are the degrees of the formed angle. Similarly, $VAL{PAR}()$ represents the value of the common slope among parallel segments and $VAL_{INT}()$ represent the location of the intersection point among the segments of the corresponding curves.

\subsection{Grammar}
Now that we have defined the alphabet for Kyrtos, we have to describe its grammar $G$. We define $G = (VN, VT, GR, F)$, where: (1) $VN$ is the set of non-terminal symbols with $VN$ =\{$Q$, $SL$, $C$, $GR$\}; (2) $VT$ is the set of terminal symbols with $VT$ = \{$V$, $@$, $!$\}; (3) $GR$ represents the starting symbol of the graphics image; (4) $F$ is the set of production rules; (5) $GR$ → [$C_i$] @ … @ [$C_j$] @ [$Q_{ijkn}$] @ ...]; (6) $Q_{ijkn}$ → [($SL_{ij}$ = $SL_{kn}$) @ ... @ ($SL_{ij+1} >< SLk_{n-2}$) @ … ]; (7) $C_i$ → [($SL_{ij}$ $\sim$ $SL_{ij+1}$ ) @ ($SL_{ij+1}$ $\sim$ $SL_{ij+2}$) @ …].
    
\subsection{Operators}
In the productions rules that we define above, we use some specific symbols. In particular: (1) $"><"$ is an operator that represents intersection among line-segments of two different curves; (2) $"="$ is an operator that represents parallelism among line-segments of two different curves; (3) $"\sim"$ is an operator that represents connection among line-segments of the same curve; (4) The $"@"$ operator represents the "and" connection among different generated relations; (5) The $"!"$ operator represents the "or" connection among different generated relations for the same curve. However, only of these relations can hold true. We use this operator when there are multiple interpretations for the same graphics image. Examples of such use case are the partitioned graphics images of the same color; (6) $[]$ and $()$ are used to determine the scope of different operations.

\section{Automatic Analysis of Charts} \label{sec_chartanalysis}

\subsection{General Analysis Methodology}
\begin{figure}[t]%% placement specifier
\centering%% For centre alignment of image.
\includegraphics[width=0.9\textwidth,height=4.3cm]{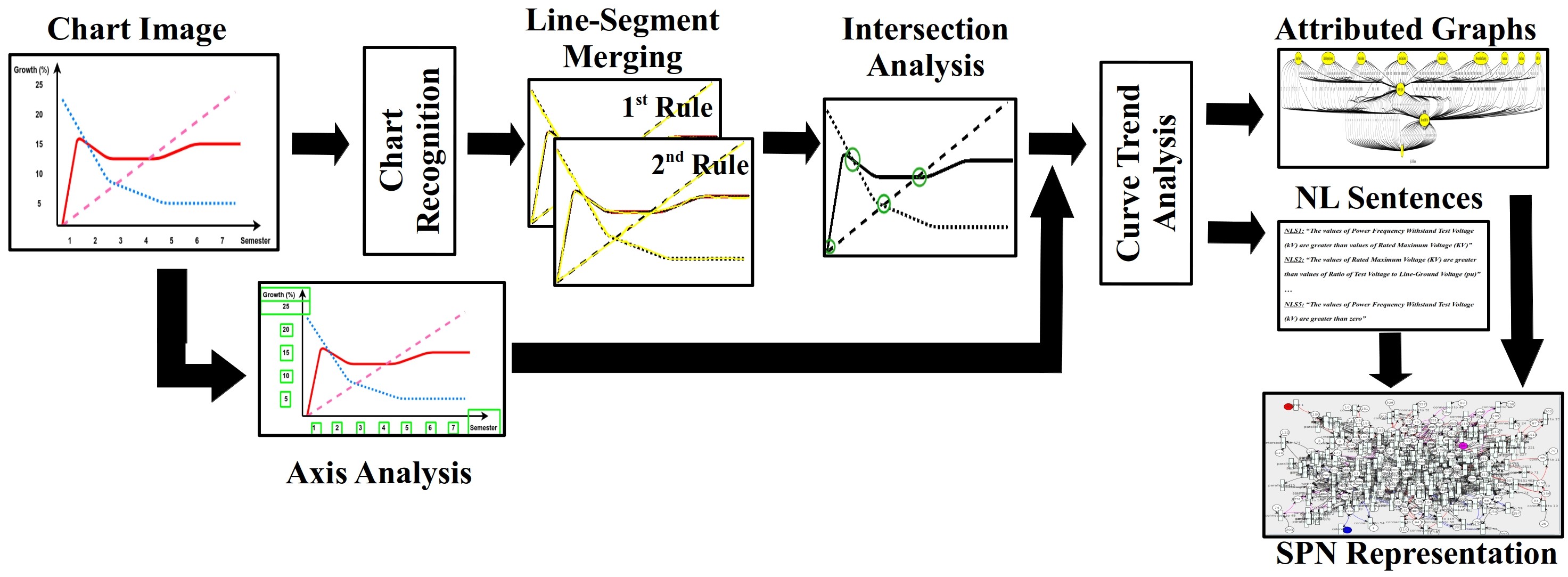}
%% Use \caption command for figure caption and label.
\caption{The Chart Analysis and Understanding Methodology of Kyrtos.}\label{fig10}
\end{figure}

It is evident in the literature review section that there have been many attempts for the processing of graphics images, however, very few research papers discuss an automated analysis methodology, and most of them work under constraining assumptions (e.g., curves with 1 or 2 pixels width). In the current section, a methodology is described for the automatic analysis of curves with arbitrary properties in chart images, based on their corresponding line-segments. The diagram for this methodology is illustrated in Fig. \ref{fig10}. To the best of our knowledge, there is no other standard method yet, that offers the same level of dynamic understanding and analysis for graphics images. Initially, the methodology condenses the recognized line-segments based on a set of heuristic merging rules to discard any excess segments. The new line-segments are analyzed based on their geometrical properties to detect regions of parallelism and intersection among different curves in the chart image. The behavioral patterns that indicate parallelism and intersections among curve segments are identified through the grammar of the Kyrtos formal language. 

The analysis methodology concludes with the extraction of information regarding growth or decay rates for each curve. More specifically, consecutive line-segments are evaluated based on their structural characteristics with respect to each curve and the resulting information are cross-correlated with the location of values in the axis to produce detailed results. In that way, the chart analysis methodology doesn’t output only the general direction but also detailed information regarding the growth of each curve in the chart image. The deduced inter- and intra-relations regarding line-segment connections, parallelisms, intersections and growth are converted into attributed graphs and natural language sentences that describe their structural properties. From those, we generate the SPN graphs that describe the functionality and behavior of each curve from the input chart image. The final SPN graph containing the structural and behavioral information of the chart’s curves is used to recreate the curves of the input chart image.

\subsection{Merging of Line Segments}

The line-segments resulting from the recognition methodology may form minor anomalies when they are connected to each other sequentially. This is due to the error in locating the middle-points with respect to each curves’ width. In the current section, two heuristic merging rules are discussed that are applied consecutively, in order to remove the unwanted anomalies formed among the segments. Both of them merge line-segments based on their geometrical attributes in correspondence to each other. During the merging procedure, we consider each line-segment as a vector instead of a simple line to avoid mathematical restrictions regarding the calculation of slope.

\begin{figure}[t]%% placement specifier
\centering%% For centre alignment of image.
\includegraphics[width=\textwidth,height=3.4cm]{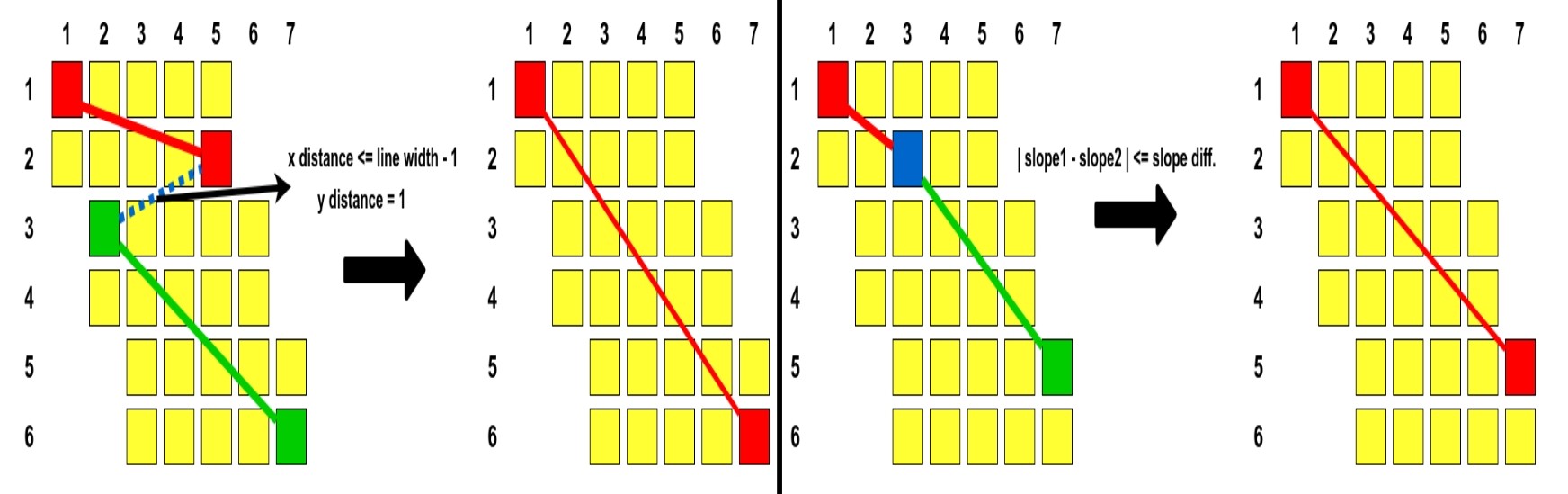}
%% Use \caption command for figure caption and label.
\caption{Left and right images show examples of the 1st and 2nd merging rules respectively.}\label{fig11}
\end{figure}

The first rule removes anomalies based on the definition of worst case lemma. \textit{Lemma: When the endpoint of the first segment has a distance of 1 pixel from the starting point of the second segment, with respect to the y-axis, and a distance of less or equal to the curve’s width minus 1 pixel, with respect to the x-axis, then we consider it as the worst case.} Any two consecutive line-segments, that fit the lemma’s description, are replaced by a larger segment with the starting point of the first and the endpoint of the second segments respectively. An illustrative example of the first merging rule is presented in Fig. \ref{fig11}. he second rule aims to discard overlapping segments that remain after the first rule. It merges two consecutive line segments into one if their slope difference is within a predefined threshold, treating each segment as a vector opposite to its connected segments. The threshold is derived by dividing the worst-case y-axis distance (which is 1) by the worst-case x-axis distance, calculated as the curve's width minus one pixel. Fig. \ref{fig11} illustrates this merging process. Together, these rules effectively preserve the curves' essential structural details.

% \begin{figure}[t]%% placement specifier
% \centering%% For centre alignment of image.
% \includegraphics[width=0.75\textwidth,height=3.5cm]{figs/fig12.png}
% %% Use \caption command for figure caption and label.
% \caption{Example of the second merging rule.}\label{fig12}
% \end{figure}

\subsection{Detection of Intersections}

In chart images where multiple curves are illustrated, these curves interact with each other through intersection. Thus, the complete understanding of structural information illustrated in graphics images depends on the computation of the intersection points among its recognized curves. In particular, these points are computed based on intersections among line-segments of different curves using Eq. \ref{eqn:eq4}, Eq. \ref{eqn:eq5}, Eq. \ref{eqn:eq6}, and Eq. \ref{eqn:eq7}. Each of the two line-segments consists one start point $(x_i, y_i)$ and one endpoint $(x_{i+1}, y_{i+1})$. We can express the position of a random point among them, as a relation of the start point $(x_1, y_1)$ plus the difference $(x_2 - x_1, y_2 - y_1)$ among them, multiplied by a variable $T$. Similarly we can express a random point for the second curve this time using variable $U$. Now, we are looking for a common point among these segments by solving the system of equations that is produced. That system of equations is solved by the program for each combination of line-segments from different curves. If the results for the variables $T$ and $U$ satisfy the conditions $0 < T < 1$ and $0 < U < 1$, then the point of intersection among these segments is computed. 

\begin{equation}\label{eqn:eq4}
(x, y) = (x_1, y_1) + T(x_2 - x_1, y_2 - y_1)
\end{equation}
\begin{equation}\label{eqn:eq5}
(x, y) = (x_3, y_3) + U(x_4 - x_3, y_4 - y_3)
\end{equation}
\begin{equation}\label{eqn:eq6}
T = [x_3  + U(x_4 - x_3) - x_1] / (x_2 - x_1)
\end{equation}
\begin{eqnarray}\label{eqn:eq7}
U= \frac{(y_4 - y_3)(x_2 - x_1) - (x_4 - x_3)(y_2 - y_1)}{y_1(x_2 - x1) - y_3(x_2 - x_1) + (x_3 - x_1)(y_2 - y_1)}
\end{eqnarray}

\subsection{Converting Curve Information to Attributed Graphs}

\begin{figure}[t]%% placement specifier
\centering%% For centre alignment of image.
\includegraphics[width=0.8\textwidth,height=3.7cm]{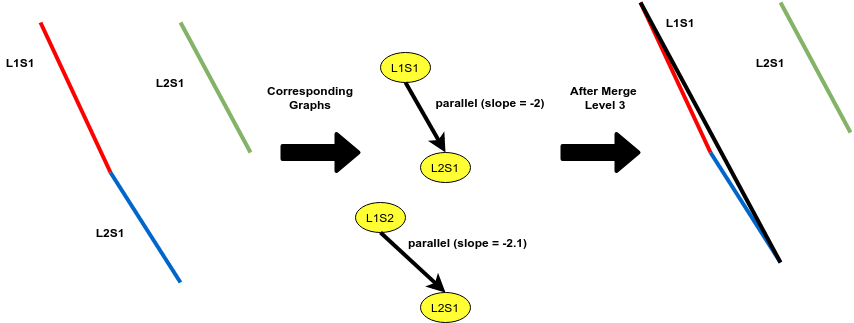}
%% Use \caption command for figure caption and label.
\caption{Example of the third merging rule.}\label{fig13}
\end{figure}

The use of attributed graphs for the representation of knowledge has already been described from Bunke et al. \cite{bunke82}, where they are used for the representation of associations and relations extracted from a circuit diagram image. More specifically, these relations correspond to the positions of circuit modules on the diagram and the connections among them. Likewise, we adopt attributed graphs to represent information extracted from a chart image, with the curves’ line-segments illustrated as nodes and the relationships among them as arcs. These relationships include information regarding connections, intersections and parallelism among different segments. Two segments from different curves are identified as parallel if they satisfy two conditions: (i) they do not intersect with each other, and (ii) their slopes are equal, allowing for a margin of difference within a threshold. This threshold is determined as the curve's identified width minus 1 pixel. Additionally, if two consecutive segments from the same curve are both parallel to a common segment from a different curve, then they are merged into a single segment (3\textsuperscript{rd} merging rule). An illustrative example of this third merging rule is presented in Fig. \ref{fig13}. 

The recognized line-segments consist the nodes of the graphs, with their respective starting and ending pixel positions represented as node attributes. The types of relationships (connection, parallelism, intersection) are represented as labels of the arc connecting two line-segments. More specifically, the connection graphs illustrate the angles of connection, measured in degrees, as their arcs’ labels. In contrast, the arcs of the intersection graphs illustrate as labels the 2D locations of intersection. Finally, parallelism graphs represent the common value of their slopes as their arcs’ labels.

\subsection{Associating the Curve Information to the Original Chart Image}

\begin{figure}[t]%% placement specifier
\centering%% For centre alignment of image.
\includegraphics[width=0.8\textwidth,height=4cm]{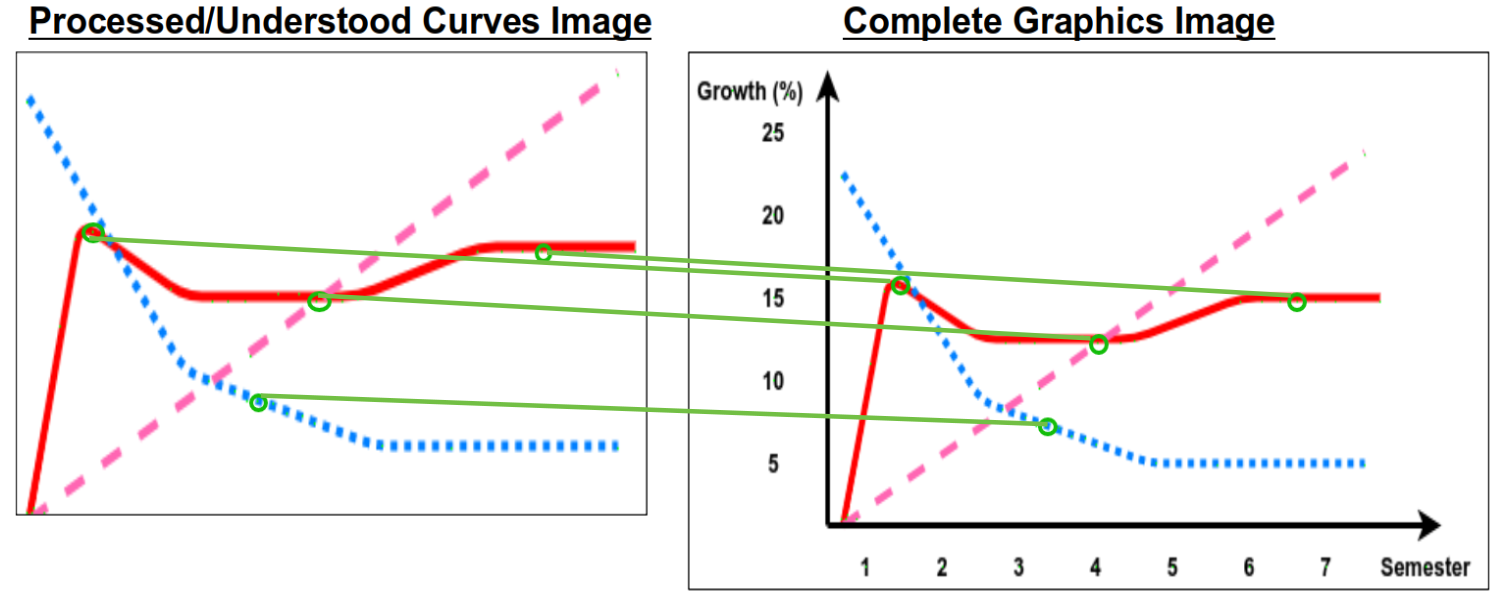}
%% Use \caption command for figure caption and label.
\caption{Example of the 2D middle-point pixel association. Left image contains the analyzed curves of the chart, right image is the original.}\label{fig15}
\end{figure}

A curve is characterized not only by its structural attributes (e.g., width, color, number of unevenness points, parallelisms) but also by higher-level behavioral information like growth rates, which indicate the increase or decrease in the curve's value over time. To recognize a curve’s growth rate, it is essential to understand the growth rates of its line segments and correlate the extracted curve information with the original chart image. This section discusses a methodology for identifying curves' growth rates by locating the midpoints of their corresponding line segments in the chart image. The approach simplifies Lowe et al. \cite{lowe}, which identifies pixel directions using gradients to extract scale and location invariant 2D points, a method effective for detecting significant points in images with high color and shape variations but less effective in images with repetitive patterns. The midpoint correlation methodology involves storing an 81x81 matrix for each midpoint, capturing the neighborhood information around these points. This matrix size is determined through trial and error and is consistently applied across all images. The position of midpoints within these matrices corresponds to their location in the curve image. For example, if a point is near the edge of the image, the matrix is adjusted accordingly. These midpoint matrices are then slid across the original chart image to find a match. The identified midpoint locations in the chart image are used to associate recognized growth rates with the values extracted from the axis. An example of this midpoint association is shown in Fig. \ref{fig15}.

For growth analysis, the curve’s line segments are assessed after applying the third merging rule, which preserves the curve’s overall direction and essential structural features. The recognition of a curve’s growth rate begins with analyzing the slopes of its consecutive line segments. If the slope of these segments is positive and continually increases, it indicates \textit{"exponential growth"}. If the slope is positive but shows little to no increase, it represents \textit{"linear growth"}. Similarly, negative slopes suggest \textit{"exponential decay"} or \textit{"linear decay"}. If the slope remains at zero across multiple segments, the curve is classified as a \textit{"steady"} line parallel to the x-axis. Further processing enables the combination of consecutive growth-rates that have the same result and, by extend the combination of their corresponding line-segments. Due to the association of the extracted middle-points with the original chart image, we are able to retrieve more detailed and enhanced knowledge about each curve’s growth-rates and, thus, achieve a deeper understanding of the chart’s underlying information.

\subsection{Converting Attributed Graphs to Natural Language Representation}

The attributed graphs are converted into natural language representation based on the information stored in the nodes and arcs of each graph relation. These NL sentences are an important transitional step for the creation of the corresponding SPN graphs. For the generation of the natural language sentences, we use the concept of Agent-Verb-Patient (AVP kernels) presented in \cite{glossa}. More specifically, the two nodes in each attributed graph relation are considered the agent (first node) and the patient (last node) of the sentence respectively. The label of the arc corresponds to the verb of the sentence.  The potential verbs are \textit{"is connected to"}, \textit{"is parallel to"}, \textit{"is intersecting with"}, \textit{"illustrates exponential/linear growth/decay"}. Certain conjunction words are adopted to enhance the semantic and grammatical integrity of the produced sentences. Examples of generated natural language sentences are the following: (a) \textit{NLS1: the straight-line segment L1S0 is connected with (34.513084935149756 degrees) to the straight-line segment L1S1}; (b) \textit{NLS109: the straight-line segment L2S3 is parallel to the straight-line segment L1S6}; (c) \textit{NLS223: the straight-line segment L2S0 is intersecting in coordinates X=111 and Y=126 with the straight-line segment L3S11}; (d) \textit{NLS316: the straight-line segment L1S2 illustrates linear growth greater than 15 \%) during 1 Semester};.

The aforementioned sentences correspond to the relations that are retrieved from the running example chart (type 1) from Fig. \ref{fig2}, after the 2\textsuperscript{nd} merging rule has been applied. Complex information such as growth-rates require more AVP kernels to be represented with NL sentences. For example, the sentence NLS316 contains 3 AVP kernels, which are \textit{"L1S1-illustrates-linear growth"}, \textit{"linear growth-greater than 15\%)"}, and \textit{"exponential growth-during-1 Semester"}.

\subsection{Converting Natural Language to SPN Representation}
SPN graphs are used to model systems with uncertainty through probabilistic distributions. They consist of (i) places representing states and (ii) transitions representing actions required to move from one state to another. Tokens flow between places and transitions, with each transition governed by a firing rate that controls progression based on available tokens, simulating the passage of time. The SPN’s ability to regulate the flow and associations of information through firing rates constitutes the distinctive feature that renders it an appealing choice for representing the extracted information of our chart analysis methodology. This control enables the progression from one state of information to another. For example, chart images illustrate the evolution of variables over time, where reaching a specific point on a curve (e.g., the highest or maximum value) requires traversing through all preceding states (i.e., prior locations on the curve). In our methodology, we emulate this concept by representing identified consecutive line-segments as places and various types of associations among them (such as connection, intersection, parallelism, growth, or decay) as transitions. In this representation, we employ the detected 2D middle-points that mark the boundaries of the line segments as tokens, serving as initiators for the progression from one line segment to the next.

This section discusses a methodology for converting NL sentences describing chart information into SPN representations, based on the methodology presented in \cite{glossa}. Each sentence is reduced to an AVP kernel (agent → verb(action) → patient), where agents and patients represent input and output places, and verbs represent transitions. These AVP kernels are then used to generate SPN graphs. The methodology is adapted to analyze NL sentences describing the behavioral characteristics of curves and the structural relationships among their line-segments using the Kyrtos formal language. The expected verbs are \textit{"is parallel to"}, \textit{"is connected to"}, or \textit{"is intersecting with"} representing associations among line segments. A transition is generated for each detected verb, while places (input or output) are generated for each line-segment based on their position in the sentence. The 2D coordinates mentioned in NL sentences act as SPN tokens, which activate transitions in the SPN graph to simulate the curve's behavior. These tokens are used only in sentences about connections and intersections, as they require locational information. The resulting SPN graphs are colored according to the recognized curve colors to highlight connections among input and output places.

\begin{figure}[t]%% placement specifier
\centering%% For centre alignment of image.
\includegraphics[width=1\textwidth,height=4.5cm]{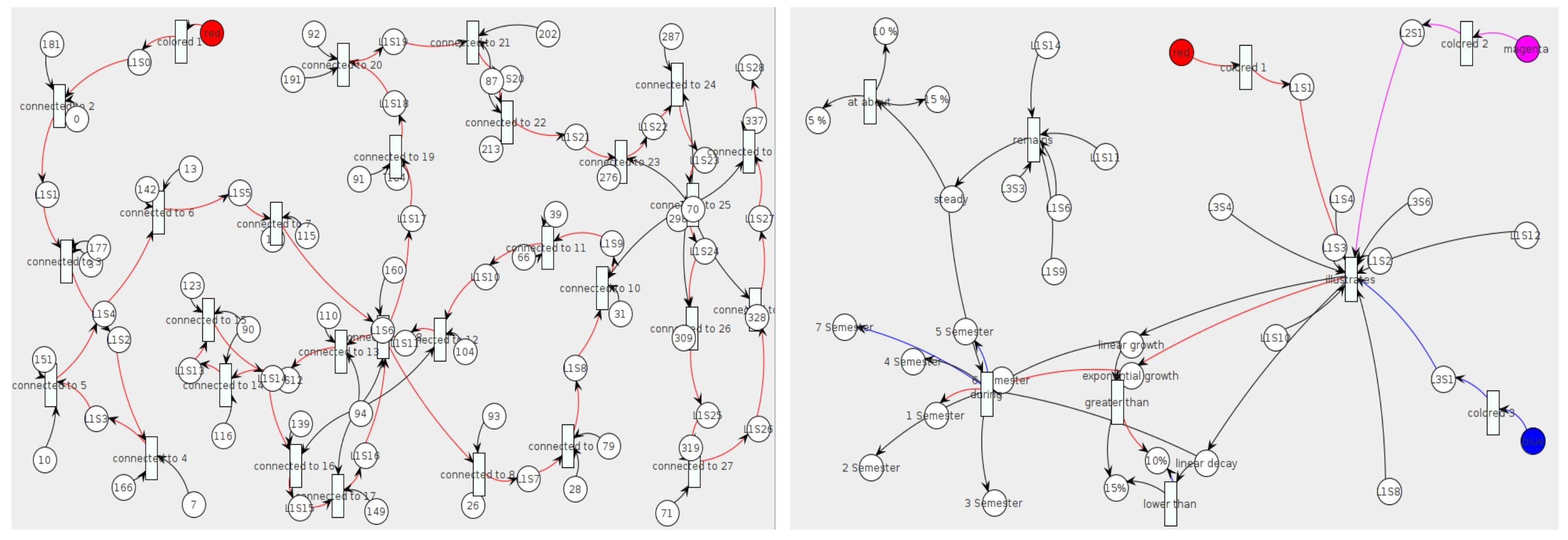}
%% Use \caption command for figure caption and label.
\caption{Left SPN graph shows the information about the red curve from the running example chart image in Fig.\ref{fig2}. The right SPN graph illustrates extracted trend information for all the curves of the running example chart image from Fig.\ref{fig2}.}\label{fig16}
\end{figure}

\begin{figure}[t]%% placement specifier
\centering%% For centre alignment of image.
\includegraphics[width=0.85\textwidth,height=5cm]{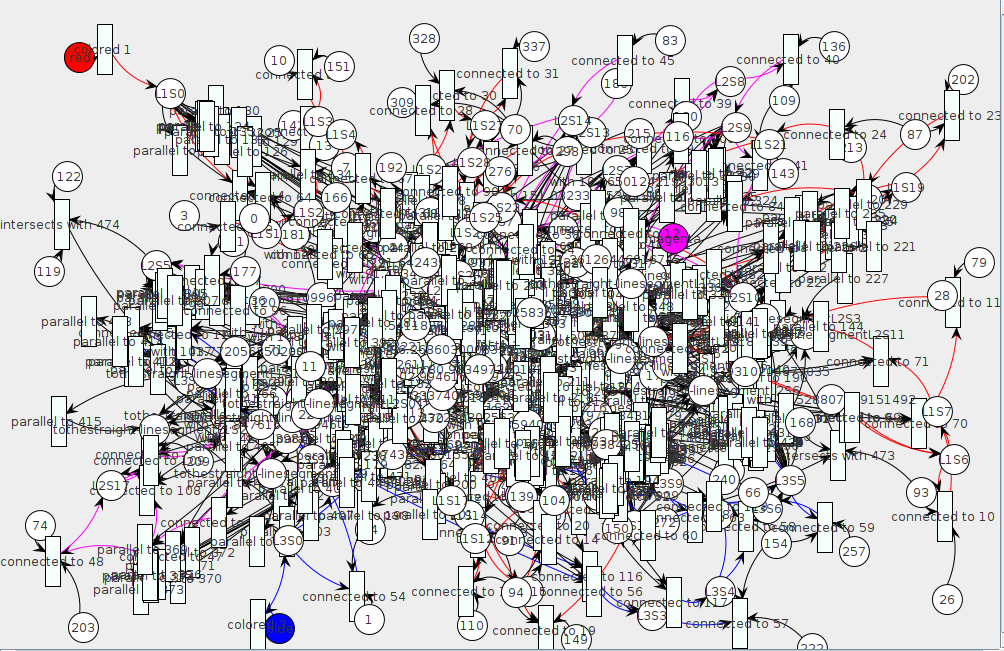}
%% Use \caption command for figure caption and label.
\caption{The SPN graph generated from all the extracted information about all curves from the running example chart image from Fig.\ref{fig2}.}\label{fig17}
\end{figure}

Fig. \ref{fig16} illustrates an example colored SPN, which contains all the connection among the line-segments of the red curve from the running example of Fig. \ref{fig2}. The $X$ and $Y$ coordinates of the starting points for each segment serve as extra tokens. An example SPN containing only the information regarding the growth-rate and trends of all the curves from the running example of Fig. \ref{fig2}, is also demonstrated in Fig. \ref{fig16}. Finally, Fig. \ref{fig17} illustrates the complete SPN output of Kyrtos, containing all the extracted structural and behavioral information about all the curves of the running example chart image from Fig. \ref{fig2}. While this final SPN graph may be complex for humans to interpret, it is fully comprehensible by a computer, aligning with the research goal of automating chart image analysis by simulating human perception of information.

\section{Evaluation of the Kyrtos Methodology based on Curve Reconstruction} \label{sec_eval}

\begin{figure}[t]%% placement specifier
\centering%% For centre alignment of image.
\includegraphics[width=\textwidth, height=3.5cm]{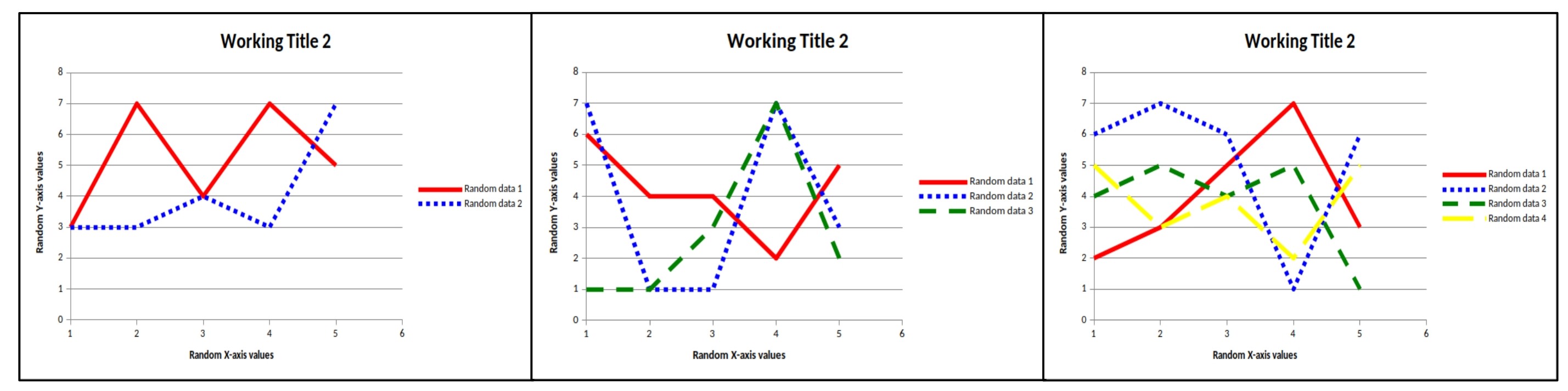}
%% Use \caption command for figure caption and label.
\caption{ Examples of charts images from the synthetic dataset for 2 curves (left), 3 curves (middle) and 4 curve (right).}\label{fig18}
\end{figure}

As part of an the initial evaluation of Kyrtos, we created a test set through Google web-crawling, consisting of chart images with single curves across four classes: linear, quadratic, asymptotic, and sinusoidal (25 images each). Existing datasets like DVQA \cite{chart_surv} mainly focus on bar charts or infographics, making them unsuitable for evaluating Kyrtos, which uses line-tracing and approximation techniques. To address this, we generated an additional synthetic dataset \footnote{https://drive.google.com/file/d/1cnkDUyqDW7FAaUYfF8cWHqhQycSEeEMA/view?usp=share\_link} comprising 150 charts with realistic curves in various directions and colors, including different dot shapes (e.g., dashed, squares) to increase difficulty. The dataset is divided into three classes: 50 charts with 2 curves, 50 with 3 curves, and 50 with 4 curves, each represented as Excel files containing both the data and the corresponding chart images. Fig. \ref{fig18} shows examples from the synthetic dataset.

The proposed Kyrtos methodology cannot be adequately evaluated against other state-of-the-art techniques, since there are no publicly available implementations of comparable methods, such as \cite{nair}, to the best of our knowledge. Therefore, we have conducted an ablation study focusing on our merging rules. Specifically, for charts containing 2 or more curves, we have applied both the 2\textsuperscript{nd} and 3\textsuperscript{rd} merging rules and evaluated the accuracy of approximation using the Structural Similarity Index Metric (SSIM). Kyrtos produces SPNs that hold both the structural (connections, parallelism and intersections) and behavioral (curve’s trend, color) information illustrated the input chart. Thus, approximations of the original curves are reconstructed based on that information. It is important to note that this evaluation approach is not applicable to single-curve charts, as the 3\textsuperscript{rd} merging rule relies on the presence of parallelism among different curves. Thus, it used only for the evaluation on the synthetic arbitrary dataset. To achieve granular evaluation and understand the efficiency of each version of the method, separate tests are performed for each category of the arbitrary chart images (2 curves, 3 curves and 4 curves).

\subsection{Qualitative Analysis}

\begin{figure}[t]%% placement specifier
\centering%% For centre alignment of image.
\includegraphics[width=\textwidth, height=5.5cm]{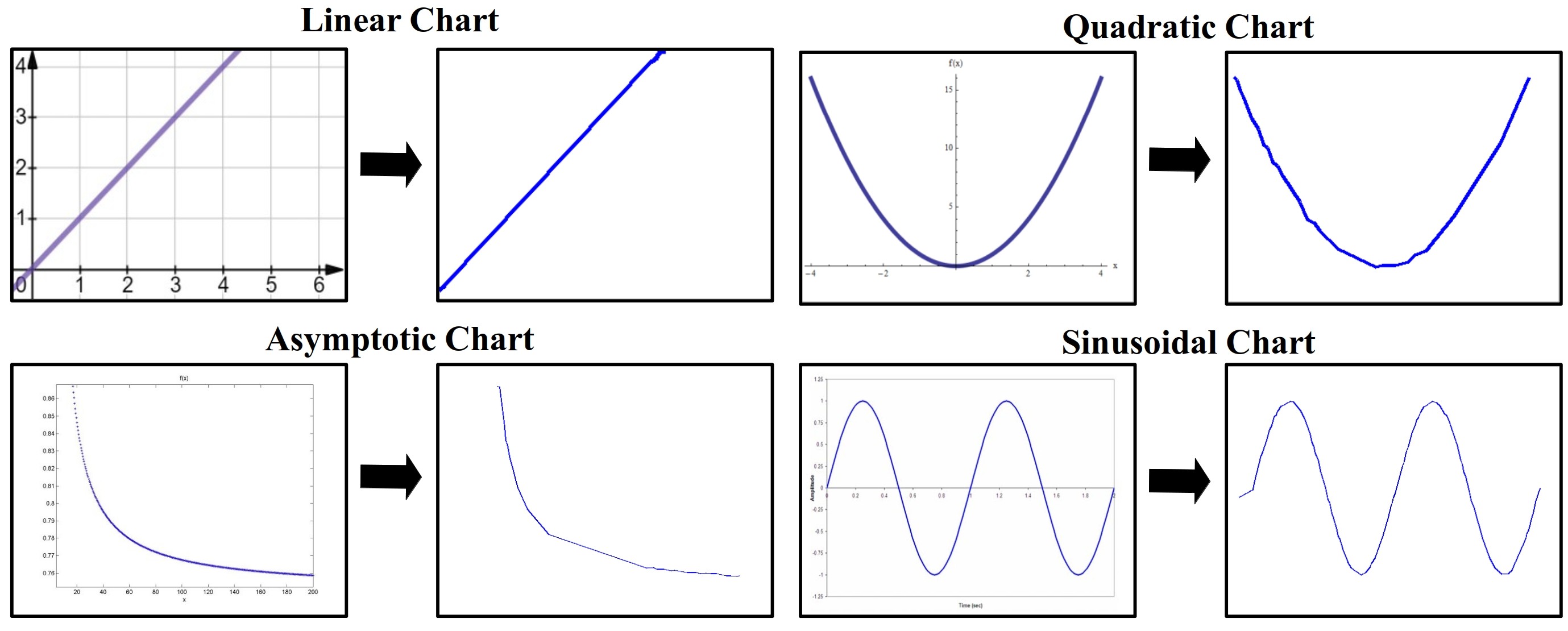}
%% Use \caption command for figure caption and label.
\caption{Example of qualitative results for different input charts from the synthetic dataset.}\label{fig19}
\end{figure}

\begin{figure}[t]%% placement specifier
\centering%% For centre alignment of image.
\includegraphics[width=0.6\textwidth, height=3cm]{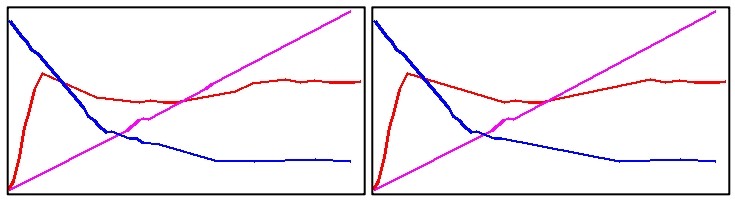}
%% Use \caption command for figure caption and label.
\caption{Example of qualitative results for different input charts from the synthetic dataset.}\label{fig20}
\end{figure}

\begin{figure}[t]%% placement specifier
\centering%% For centre alignment of image.
\includegraphics[width=\textwidth, height=9cm]{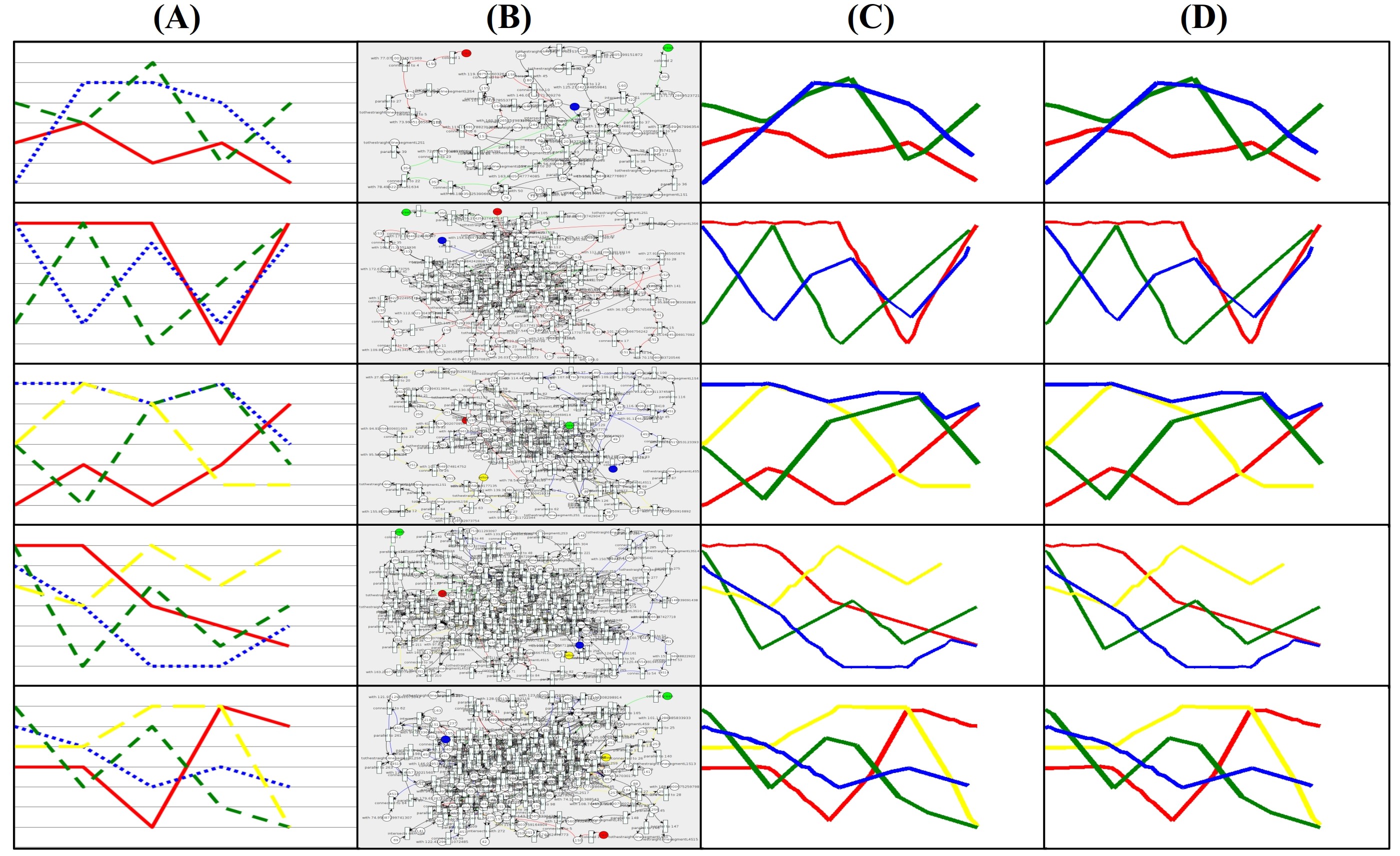}
%% Use \caption command for figure caption and label.
\caption{Example of qualitative results for different input charts from the synthetic dataset.}\label{fig21}
\end{figure}

This subsection demonstrates Kyrtos' performance by visualizing the reconstruction results for various chart classes from the synthetic dataset. Fig. \ref{fig19} shows Kyrtos’ reconstruction for linear, asymptotic, quadratic, and sinusoidal curves. Fig. \ref{fig20} presents results from the running example in Fig. \ref{fig2}, comparing reconstructions using the 2\textsuperscript{nd} merging rule (left) and both the 2\textsuperscript{nd} and 3\textsuperscript{rd} merging rules (right). Fig. \ref{fig21} highlights reconstruction results for five charts, showing input subimages (A), SPN graphs after all merging rules (B), and reconstructions using the 2\textsuperscript{nd} (C) and all merging rules (D). It is worth noting that the reconstruction results in column C exhibit a closer resemblance to the original curves compared to those in column D. This outcome is expected since the purpose of the 3\textsuperscript{rd} merging rule is to remove consecutive segments with similar traits by approximating their behavior with a single segment. Consequently, this enhances computational efficiency by reducing the number of segments for analysis. However, there is a trade-off between computational speed and granularity/accuracy, as the use of the 3\textsuperscript{rd} merging rule sacrifices some level of detail.

The results of Fig. \ref{fig21} also demonstrate the limitations of the proposed method, revealing instances where Kyrtos struggles to correctly identify the endpoints of the blue curves in the 3\textsuperscript{rd} and 4\textsuperscript{th} rows. After conducting extensive experimentation, we have identified that this specific error is associated with inaccuracies during the removal of axis pixels. These persistent axis pixels are located near the endpoints of other curves and, thus, also manage to evade detection by the noise-pixel filter. Furthermore, in the 1st row of Fig. \ref{fig21},  it is evident that Kyrtos does not accurately recognize the height of the curves' top corners. The proposed method approximates the structural and behavioral characteristics of a curve by employing clustering techniques on sets of selected unevenness keypoints. However, these keypoints are extracted using a heuristic line-tracing algorithm, which may prioritize lower-corner keypoints over top-corner keypoints due to the parsing order. Consequently, fewer unevenness keypoints are considered during the clustering phase, limiting the accuracy of the approximation.

\subsection{Quantitative Analysis}

In \cite{chart_surv}, natural language-based evaluation techniques are discussed, primarily for chart visual question answering and text recognition tasks that generate textual outputs. In contrast, Kyrtos focuses on analyzing chart curves using line tracing and approximation techniques, necessitating different evaluation methods. Therefore, to assess Kyrtos' effectiveness, we measure the distance between approximated 2D middle-points and their actual locations. However, due to the difficulty in precisely determining the actual location of middle-points, we instead reconstruct the curves using the approximated values and apply the SSIM metric to objectively evaluate accuracy again the original curves. This approach avoids the subjectivity inherent in human grading and aligns with methods in \cite{chart_surv} and \cite{nair}, where reconstruction is also used for evaluation.

\begin{table}[t]%% placement specifier
%% Use tabular environment to tag the tabular data.
\caption{Curve reconstruction evaluation using structural similarity for single-curve dataset.}
\label{tbl1}
\centering%% For centre alignment of tabular.
\begin{tabular}{l c r}%% Table column specifiers
%% Tabular cells are separated by &
\hline
  \textbf{Category of Charts} & \textbf{SSIM} & \textbf{Standard Deviation}\\ %% A tabular row ends with \\
\hline
  Linear & 0.96 & 0.000012 \\
  Quadratic & 0.95 & 0.00094 \\
  Asymptotic & 0.97 & 0.00024\\
  Sinusoidal & 0.97 & 0.00015\\
\hline
\end{tabular}
\end{table}

Table \ref{tbl1} illustrates the results from evaluating Kyrtos’ accuracy for the reconstruction of the charts containing linear, quadratic, asymptotic and sinusoidal. It is evident from these results, that in all cases, the recognition and analysis methods of Kyrtos produce highly accurate approximations. Table \ref{tbl2} presents the reconstruction results achieved by two versions of Kyrtos: one utilizing only the 1\textsuperscript{st} and 2\textsuperscript{nd} merging rules, and the other employing all three merging rules, across varying numbers of arbitrary curves in the chart. The resulting SSIM accuracy scores align with the performed qualitative analysis, as the version of Kyrtos that employed the 2\textsuperscript{nd} merging rule consistently exhibits higher accuracy for all numbers of curves compared to its counterpart. Additionally, we observe a gradual decrease in accuracy across all methods as the number of curves in the input chart image increases. This observation can be attributed to the limitations of the unevenness keypoint extraction method employed. Specifically, as a chart image contains a larger number of curves, it tends to exhibit more intricate details and information. However, the performance of the clustering-based keypoint selection methods deteriorates as the number of curves in the chart increases, showcasing the necessity for a more adaptable and robust approach to handle such complex scenarios.

\begin{table}[t]%% placement specifier
%% Use tabular environment to tag the tabular data.
\caption{Curve reconstruction evaluation using structural similarity for synthetic arbitrary dataset.}
\label{tbl2}
\centering%% For centre alignment of tabular.
\begin{tabular}{l c r}%% Table column specifiers
%% Tabular cells are separated by &
\hline
  \textbf{Category of Charts} & \textbf{SSIM} & \textbf{Standard Deviation}\\ %% A tabular row ends with \\
\hline
  Arbitrary 2 Curves (2\textsuperscript{nd} Merging Rule) & 0.92 & 0.0018 \\
  Arbitrary 2 Curves (2\textsuperscript{nd} \& 3\textsuperscript{rd} Merging Rules) & 0.91 & 0.0043 \\
  Arbitrary 3 Curves (2\textsuperscript{nd} Merging Rule) & 0.86 & 0.0031\\
  Arbitrary 3 Curves (2\textsuperscript{nd} \& 3\textsuperscript{rd} Merging Rules) & 0.84 & 0.0057\\
  Arbitrary 4 Curves (2\textsuperscript{nd} Merging Rule) & 0.82 & 0.0076\\
  Arbitrary 4 Curves (2\textsuperscript{nd} \& 3\textsuperscript{rd} Merging Rules) & 0.79 & 0.0053\\
\hline
\end{tabular}
\end{table}

\section{Conclusions \& Future Work}\label{sec_conclude}
In conclusion, we have presented Kyrtos, a methodology for automatically recognizing and analyzing chart images from technical documents, converting their structural and behavioral information into SPN graphs. Unlike other methods that rely on heuristic restrictions or handle only simple charts, Kyrtos handles images with dynamic and complex features. The method identifies 2D middle-points to reconstruct the curves in a chart, using these points to extract structural (e.g., connections, intersections) and behavioral (e.g., direction, trend) information. This information is then converted into an SPN graph for accurate curve reconstruction. Kyrtos provides approximations of input curves with high accuracy, due to being invariant to color, width and directional information. More specifically, the application of the HSV homogeneity filter enables the coloring of pixels with different shades of the same color, with their dominant color value, thus reducing the amount of noise in the chart image without discarding any important information. Similarly, the application of an hierarchical clustering-based filter during recognition, exploits the curve’s width for the selection of its representative 2D middle-points.

Despite it’s accurate results in the recognition and analysis of chart information, the discussed methodology is limited to charts containing curves of different colors. Therefore, its accuracy will drop significantly in cases where all the curves in the chart image are illustrated using the same color, even if they have different structural characteristics (e.g., lines with dots or dashes). Furthermore, Kyrtos constitutes a computationally intensive methodology, since its recognition procedure requires a lot of time to extract accurately the 2D middle-points. Its complexity depends on the resolution of the input chart image, the number of differently colored curves, as well as the number of values on the axis. Further study is required to expand the current implementation of Kyrtos, to analyze bar charts and charts containing curves of the same color. For the former, it will be crucial to develop a different keypoint extraction method, since the current version of Kyrtos relies on line-tracing and approximation techniques for curve analysis. On the contrary, processing charts containing curves of the same color would require an additional statistical model to complement the existing curve approximation approach, enabling the resolution of conflicts arising from shared pixels of different curves. To accommodate these charts, the NL representation kernels and the SPN graph representations should be updated to reflect the additional relationships and concepts, such as column-to-column ratio in bar charts and the uncertainty of keypoint origin in charts with curves of the same color.

The overall objective of this research is to develop a document analysis methodology inspired by human perception. Humans can intuitively connect information across different modalities within a document and understand the conditional relationships between events expressed visually or textually. To emulate this, our document analysis methodology will independently analyze each modality, converting the information into a unified representation that integrates seamlessly while preserving temporal relationships. This approach enhances our understanding of the document's underlying connections. Similar methods for extracting and analyzing information from tables and diagrams are discussed in \cite{pinakas}, and we plan to merge their SPN graphs synergistically. Additionally, Kyrtos is methodology focused on achieving high accuracy and doesn’t constitute a real-time solution, due its to slow recognition procedure. Inspired by the work presented in \cite{rodeo}, where a methodology is discussed for the reconstruction of medical images from sampled data, we plan to train an Autoencoder to extract the 2D data-points that are sufficient for the reconstruction of curves. More specifically, an initial dataset can be built based on the outputs of Kyrtos, containing the 2D keypoints extracted from various charts images. Finally, we plan to study the effectiveness of Kyrtos in damaged documents \cite{tanlu21}. Since Kyrtos is invariant to curve segmentations, it could assist with enhancing chart information by approximating chart curves.

\section*{Acknowledgements}
This work is partially supported by an ONR-BAA grant 2018-2022.





% %% For numbered reference style

%   \textit{\LaTeX: a document preparation system},
%   Addison Wesley, Massachusetts,

% \end{thebibliography}

\bibliographystyle{elsarticle-num}
\bibliography{sample}

@article{biblio_survey,
title = {A bibliometric analysis of off-line handwritten document analysis literature (1990–2020)},
journal = {Pattern Recognition},
volume = {125},
pages = {108513},
year = {2022},
issn = {0031-3203},
doi = {https://doi.org/10.1016/j.patcog.2021.108513},
author = {Victoria Ruiz-Parrado and Ruben Heradio and Ernesto Aranda-Escolastico and Ángel Sánchez and José F. Vélez}
}

@article{bunke11,
title = {Recent advances in graph-based pattern recognition with applications in document analysis},
journal = {Pattern Recognition},
volume = {44},
number = {5},
pages = {1057-1067},
year = {2011},
issn = {0031-3203},
doi = {https://doi.org/10.1016/j.patcog.2010.11.015},
author = {Horst Bunke and Kaspar Riesen}
}

@INPROCEEDINGS{cao23,
  author={Cao, Haoyu and Bao, Changcun and Liu, Chaohu and Chen, Huang and Yin, Kun and Liu, Hao and Liu, Yinsong and Jiang, Deqiang and Sun, Xing},
  booktitle={2023 IEEE/CVF International Conference on Computer Vision (ICCV)}, 
  title={Attention Where It Matters: Rethinking Visual Document Understanding with Selective Region Concentration}, 
  year={2023},
  volume={},
  number={},
  pages={19460-19470},
  doi={10.1109/ICCV51070.2023.01788}}

@InProceedings{cvprli24,
    author    = {Li, Xin and Wu, Yunfei and Jiang, Xinghua and Guo, Zhihao and Gong, Mingming and Cao, Haoyu and Liu, Yinsong and Jiang, Deqiang and Sun, Xing},
    title     = {Enhancing Visual Document Understanding with Contrastive Learning in Large Visual-Language Models},
    booktitle = {Proceedings of the IEEE/CVF Conference on Computer Vision and Pattern Recognition (CVPR)},
    month     = {June},
    year      = {2024},
    pages     = {15546-15555}
}

@article{gordo13,
title = {Large-scale document image retrieval and classification with runlength histograms and binary embeddings},
journal = {Pattern Recognition},
volume = {46},
number = {7},
pages = {1898-1905},
year = {2013},
issn = {0031-3203},
doi = {https://doi.org/10.1016/j.patcog.2012.12.004},
author = {Albert Gordo and Florent Perronnin and Ernest Valveny}
}

@article{pinakas,
author = {Alexiou, Michail S. and Bourbakis, Nikolaos G.},
title = {Pinakas: A Methodology for Deep Analysis of Tables in Technical Documents},
journal = {International Journal on Artificial Intelligence Tools},
volume = {32},
number = {04},
pages = {2350042},
year = {2023},
doi = {10.1142/S0218213023500422}
}

@article{rodeo,
title = {RODEO: Robust DE-aliasing autoencOder for real-time medical image reconstruction},
journal = {Pattern Recognition},
volume = {63},
pages = {499-510},
year = {2017},
issn = {0031-3203},
doi = {https://doi.org/10.1016/j.patcog.2016.09.022},
author = {Janki Mehta and Angshul Majumdar}
}

@article{selosse,
title = {Textual data summarization using the Self-Organized Co-Clustering model},
journal = {Pattern Recognition},
volume = {103},
pages = {107315},
year = {2020},
issn = {0031-3203},
doi = {https://doi.org/10.1016/j.patcog.2020.107315},
author = {Margot Selosse and Julien Jacques and Christophe Biernacki}
}

@article{riba22,
title = {Table detection in business document images by message passing networks},
journal = {Pattern Recognition},
volume = {127},
pages = {108641},
year = {2022},
issn = {0031-3203},
doi = {https://doi.org/10.1016/j.patcog.2022.108641},
author = {Pau Riba and Lutz Goldmann and Oriol Ramos Terrades and Diede Rusticus and Alicia Fornés and Josep Lladós}
}

@article{tanlu21,
title = {Probabilistic homogeneity for document image segmentation},
journal = {Pattern Recognition},
volume = {109},
pages = {107591},
year = {2021},
issn = {0031-3203},
doi = {https://doi.org/10.1016/j.patcog.2020.107591},
author = {Tan Lu and Ann Dooms}
}

@article{passalis18,
title = {Learning bag-of-embedded-words representations for textual information retrieval},
journal = {Pattern Recognition},
volume = {81},
pages = {254-267},
year = {2018},
issn = {0031-3203},
doi = {https://doi.org/10.1016/j.patcog.2018.04.008},
author = {Nikolaos Passalis and Anastasios Tefas}
}

@article{chaieb17,
title = {Fuzzy generalized median graphs computation: Application to content-based document retrieval},
journal = {Pattern Recognition},
volume = {72},
pages = {266-284},
year = {2017},
issn = {0031-3203},
doi = {https://doi.org/10.1016/j.patcog.2017.07.030},
author = {Ramzi Chaieb and Karim Kalti and Muhammad Muzzamil Luqman and Mickaël Coustaty and Jean-Marc Ogier and Najoua {Essoukri Ben Amara}}
}

@ARTICLE{chart_surv,
  author={Davila, Kenny and Setlur, Srirangaraj and Doermann, David and Kota, Bhargava Urala and Govindaraju, Venu},
  journal={IEEE Transactions on Pattern Analysis and Machine Intelligence}, 
  title={Chart Mining: A Survey of Methods for Automated Chart Analysis}, 
  year={2021},
  volume={43},
  number={11},
  pages={3799-3819},
  doi={10.1109/TPAMI.2020.2992028}}

@INPROCEEDINGS{graph_surv,
  author={Alexiou, Michail S. and Bourbakis, Nikolaos G.},
  booktitle={2021 IEEE 33rd International Conference on Tools with Artificial Intelligence (ICTAI)}, 
  title={An Evaluation of Methods on Detecting, Recognizing and Understanding Graphics Images in Technical Documents}, 
  year={2021},
  volume={},
  number={},
  pages={568-574},
  doi={10.1109/ICTAI52525.2021.00091}}

@inproceedings{greenbacker,
author = {Greenbacker, Charles F. and Wu, Peng and Carberry, Sandra and McCoy, Kathleen F. and Elzer, Stephanie},
title = {Abstractive summarization of line graphs from popular media},
year = {2011},
isbn = {9781932432947},
publisher = {Association for Computational Linguistics},
address = {USA},
booktitle = {Proceedings of the Workshop on Automatic Summarization for Different Genres, Media, and Languages},
pages = {41–48},
doi = {10.5555/2018987.2018993},
numpages = {8},
location = {Portland, Oregon},
series = {WASDGML '11}
}

@article{burns,
author = {Burns, Richard and Carberry, Sandra and Elzer Schwartz, Stephanie},
title = {An automated approach for the recognition of intended messages in grouped bar charts},
journal = {Computational Intelligence},
volume = {35},
number = {4},
pages = {955-1002},
doi = {https://doi.org/10.1111/coin.12227},
year = {2019}
}

@InProceedings{wupeng1,
author="Wu, Peng
and Carberry, Sandra
and Elzer, Stephanie
and Chester, Daniel",
editor="Goel, Ashok K.
and Jamnik, Mateja
and Narayanan, N. Hari",
title="Recognizing the Intended Message of Line Graphs ",
booktitle="Diagrammatic Representation and Inference",
year="2010",
publisher="Springer Berlin Heidelberg",
address="Berlin, Heidelberg",
pages="220--234",
isbn="978-3-642-14600-8"
}

@inproceedings{wupeng2,
author = {Wu, Peng and Carberry, Sandra and Elzer, Stephanie},
title = {Segmenting Line Graphs into Trends},
booktitle = {Proceedings of the Twelth International Conference on Artificial Intelligence},
pages = {697-703},
doi = {https://doi.org/10.1111/coin.12227},
year = {2010}
}

@inproceedings{carberry,
author = {Carberry, Sandra and Elzer, Stephanie and Demir, Seniz},
title = {Information graphics: an untapped resource for digital libraries},
year = {2006},
isbn = {1595933697},
publisher = {Association for Computing Machinery},
address = {New York, NY, USA},
doi = {10.1145/1148170.1148270},
booktitle = {Proceedings of the 29th Annual International ACM SIGIR Conference on Research and Development in Information Retrieval},
pages = {581–588},
numpages = {8},
location = {Seattle, Washington, USA},
series = {SIGIR '06}
}

@inproceedings{chester,
author = {Chester, Daniel and Elzer, Stephanie},
title = {Getting computers to see information graphics so users do not have to},
year = {2005},
isbn = {3540258787},
publisher = {Springer-Verlag},
address = {Berlin, Heidelberg},
doi = {10.1007/11425274\_68},
booktitle = {Proceedings of the 15th International Conference on Foundations of Intelligent Systems},
pages = {660–668},
numpages = {9},
location = {Saratoga Springs, NY},
series = {ISMIS'05}
}

@INPROCEEDINGS{alkady,
  author={Al-Kady, Yasmin and Abdel-Wahab, Nurhan and Hassanein, Farah and Al-Alfy, Rana and El-Malatawy, Lama and Ammar, Rowym and Al-Shetairy, Mirna and Amin, Salsabil},
  booktitle={2023 Eleventh International Conference on Intelligent Computing and Information Systems (ICICIS)}, 
  title={Chartlytics: Chart Image Classification and Data Extraction}, 
  year={2023},
  volume={},
  number={},
  pages={515-522},
  doi={10.1109/ICICIS58388.2023.10391160}}

@ARTICLE{dou24,
  author={Dou, Shuguang and Jiang, Xinyang and Liu, Lu and Ying, Lu and Shan, Caihua and Shen, Yifei and Dong, Xuanyi and Wang, Yun and Li, Dongsheng and Zhao, Cairong},
  journal={IEEE Transactions on Pattern Analysis and Machine Intelligence}, 
  title={Hierarchical Recognizing Vector Graphics and A New Chart-based Vector Graphics Dataset}, 
  year={2024},
  volume={},
  number={},
  pages={1-18},
  doi={10.1109/TPAMI.2024.3394298}}

@article{kataria1,
  title={Automated analysis of images in documents for intelligent document search},
  author={Xiaonan Lu and Saurabh Kataria and William Browuer and James Ze Wang and Prasenjit Mitra and C. Lee Giles},
  journal={International Journal on Document Analysis and Recognition (IJDAR)},
  year={2009},
  volume={12},
  pages={65-81},
  doi={https://doi.org/10.1007/s10032-009-0081-0}
}

@InProceedings{cliche,
author="Cliche, Mathieu
and Rosenberg, David
and Madeka, Dhruv
and Yee, Connie",
editor="Ceci, Michelangelo
and Hollm{\'e}n, Jaakko
and Todorovski, Ljup{\v{c}}o
and Vens, Celine
and D{\v{z}}eroski, Sa{\v{s}}o",
title="Scatteract: Automated Extraction of Data from Scatter Plots",
booktitle="Machine Learning and Knowledge Discovery in Databases",
year="2017",
publisher="Springer International Publishing",
address="Cham",
pages="135--150",
isbn="978-3-319-71249-9"
}

@article{alzaidy, title={A Machine Learning Approach for Semantic Structuring of Scientific Charts in Scholarly Documents}, volume={31}, DOI={10.1609/aaai.v31i2.19088}, abstractNote={&lt;p&gt;Large scholarly repositories are designed to provide scientists and researchers with a wealth of information that is retrieved from data present in a variety of formats. A typical scholarly document contains information in a combined layout of texts and graphic images. Common types of graphics found in these documents are scientific charts that are used to represent data values in a visual format. Experimental results are rarely described without the aid of one form of a chart or another, whether it is 2D plot, bar chart, pie chart, etc. Metadata of these graphics is usually the only content that is made available for search by user queries. By processing the image content and extracting the data represented in the graphics, search engines will be able to handle more specific queries related to the data itself. In this paper we describe a machine learning based system that extracts and recognizes the various data fields present in a bar chart for semantic labeling. Our approach comprises of a graphics and text separation and extraction phase, followed by a component role classification for both text and graphic components that are in turn used for semantic analysis and representation of the chart. The proposed system is tested on a set of over 200 bar charts extracted from over 1,000 scientific articles in PDF format.&lt;/p&gt;}, number={2}, journal={Proceedings of the AAAI Conference on Artificial Intelligence}, author={Al-Zaidy, Rabah and Giles, C.}, year={2017}, month={Feb.}, pages={4644-4649} }

@INPROCEEDINGS{nair,
  author={Nair, Rathin Radhakrishnan and Sankaran, Nishant and Nwogu, Ifeoma and Govindaraju, Venu},
  booktitle={2015 13th International Conference on Document Analysis and Recognition (ICDAR)}, 
  title={Automated analysis of line plots in documents}, 
  year={2015},
  volume={},
  number={},
  pages={796-800},
  doi={10.1109/ICDAR.2015.7333871}}

@inproceedings{kataria2,
author = {Kataria, Saurabh and Browuer, William and Mitra, Prasenjit and Giles, C. Lee},
title = {Automatic extraction of data points and text blocks from 2-dimensional plots in digital documents},
year = {2008},
isbn = {9781577353683},
publisher = {AAAI Press},
booktitle = {Proceedings of the 23rd National Conference on Artificial Intelligence - Volume 2},
pages = {1169–1174},
numpages = {6},
location = {Chicago, Illinois},
series = {AAAI'08}
}

@inproceedings{kim4,
author = {Kim, Edward and McCoy, Kathleen F.},
title = {Multimodal Deep Learning using Images and Text for Information Graphic Classification},
year = {2018},
isbn = {9781450356503},
publisher = {Association for Computing Machinery},
address = {New York, NY, USA},
doi = {10.1145/3234695.3236357},
booktitle = {Proceedings of the 20th International ACM SIGACCESS Conference on Computers and Accessibility},
pages = {143–148},
numpages = {6},
location = {Galway, Ireland},
series = {ASSETS '18}
}

@INPROCEEDINGS{li23,
  author={Li, Yazhou and Ma, Yanyun and Dai, Wei and Zhang, Weifang},
  booktitle={2023 IEEE International Conference on Industrial Engineering and Engineering Management (IEEM)}, 
  title={Control Chart Pattern Recognition Based on MDWOP and Ensemble Classifier}, 
  year={2023},
  volume={},
  number={},
  pages={1763-1767},
  doi={10.1109/IEEM58616.2023.10406686}}

@article{unevennes,
title = {Recognition of line segments with unevenness used in OCR and fingerprints},
journal = {Engineering Applications of Artificial Intelligence},
volume = {12},
number = {3},
pages = {273-279},
year = {1999},
issn = {0952-1976},
doi = {https://doi.org/10.1016/S0952-1976(98)00063-3},
author = {Nikolaos G Bourbakis and David Goldman}
}

@ARTICLE{bunke82,
  author={Bunke, Horst},
  journal={IEEE Transactions on Pattern Analysis and Machine Intelligence}, 
  title={Attributed Programmed Graph Grammars and Their Application to Schematic Diagram Interpretation}, 
  year={1982},
  volume={PAMI-4},
  number={6},
  pages={574-582},
  doi={10.1109/TPAMI.1982.4767310}}

@INPROCEEDINGS{lowe,
  author={Lowe, D.G.},
  booktitle={Proceedings of the Seventh IEEE International Conference on Computer Vision}, 
  title={Object recognition from local scale-invariant features}, 
  year={1999},
  volume={2},
  number={},
  pages={1150-1157 vol.2},
  doi={10.1109/ICCV.1999.790410}}

@INPROCEEDINGS {chart_reader,
author = {Z. Cheng and Q. Dai and A. G. Hauptmann},
booktitle = {2023 IEEE/CVF International Conference on Computer Vision (ICCV)},
title = {ChartReader: A Unified Framework for Chart Derendering and Comprehension without Heuristic Rules},
year = {2023},
volume = {},
issn = {},
pages = {22145-22156},
doi = {10.1109/ICCV51070.2023.02029},
publisher = {IEEE Computer Society},
address = {Los Alamitos, CA, USA},
month = {oct}
}

@InProceedings{chart_vqa,
    author    = {Li, Zhuowan and Jasani, Bhavan and Tang, Peng and Ghadar, Shabnam},
    title     = {Synthesize Step-by-Step: Tools Templates and LLMs as Data Generators for Reasoning-Based Chart VQA},
    booktitle = {Proceedings of the IEEE/CVF Conference on Computer Vision and Pattern Recognition (CVPR)},
    month     = {June},
    year      = {2024},
    pages     = {13613-13623}
}

@article{alexiou_chart,
title = {Behavioral analysis of bar charts in documents via stochastic petri-net modeling},
journal = {Pattern Recognition Letters},
volume = {176},
pages = {174-181},
year = {2023},
issn = {0167-8655},
doi = {https://doi.org/10.1016/j.patrec.2023.11.004},
author = {Michail S. Alexiou and Nikolaos G. Bourbakis}
}

@InProceedings{layoutllm,
    author    = {Luo, Chuwei and Shen, Yufan and Zhu, Zhaoqing and Zheng, Qi and Yu, Zhi and Yao, Cong},
    title     = {LayoutLLM: Layout Instruction Tuning with Large Language Models for Document Understanding},
    booktitle = {Proceedings of the IEEE/CVF Conference on Computer Vision and Pattern Recognition (CVPR)},
    month     = {June},
    year      = {2024},
    pages     = {15630-15640}
}

@INPROCEEDINGS{unifyingmodal,
  author={Tang, Zineng and Yang, Ziyi and Wang, Guoxin and Fang, Yuwei and Liu, Yang and Zhu, Chenguang and Zeng, Michael and Zhang, Cha and Bansal, Mohit},
  booktitle={2023 IEEE/CVF Conference on Computer Vision and Pattern Recognition (CVPR)}, 
  title={Unifying Vision, Text, and Layout for Universal Document Processing}, 
  year={2023},
  volume={},
  number={},
  pages={19254-19264},
  doi={10.1109/CVPR52729.2023.01845}}

@article{glossa,
author = {Psarologou, Adamantia and Bourbakis, Nikolaos},
title = {Glossa — A Formal Language as a Mapping Mechanism of NL Sentences into SPN State Machine for Actions/Events Association},
journal = {International Journal on Artificial Intelligence Tools},
volume = {26},
number = {02},
pages = {1750012},
year = {2017},
doi = {10.1142/S0218213017500129}
}

@InProceedings{doc_assist,
    author    = {Liu, Chaohu and Yin, Kun and Cao, Haoyu and Jiang, Xinghua and Li, Xin and Liu, Yinsong and Jiang, Deqiang and Sun, Xing and Xu, Linli},
    title     = {HRVDA: High-Resolution Visual Document Assistant},
    booktitle = {Proceedings of the IEEE/CVF Conference on Computer Vision and Pattern Recognition (CVPR)},
    month     = {June},
    year      = {2024},
    pages     = {15534-15545}
}

\end{document}